\documentclass[letterpaper]{article} 
\usepackage{aaai2026}  
\nocopyright 
\usepackage{times}  
\usepackage{helvet}  
\usepackage{courier}  
\usepackage[hyphens]{url}  
\usepackage{graphicx} 
\usepackage{natbib}  
\usepackage{multirow}
\usepackage{array}
\usepackage{longtable}
\usepackage{float}
\usepackage{subcaption}  
\usepackage{caption} 
\usepackage{algorithm}
\usepackage{algorithmic}

\usepackage{amsmath,amsfonts,amssymb}
\usepackage{booktabs}
\usepackage{mathtools}
\usepackage{bm}
\usepackage{fancyhdr}
\title{CrystalDiT: A Diffusion Transformer for Crystal Generation}

\author{
    Xiaohan Yi\textsuperscript{1,2}, 
    Guikun Xu\textsuperscript{3},    
    Zhong Zhang\textsuperscript{2},  
    Liu Liu\textsuperscript{2},      
    Yatao Bian\textsuperscript{4},   
    Xi Xiao\textsuperscript{1}\thanks{Corresponding author.}, 
    Peilin Zhao\textsuperscript{3}\thanks{Corresponding author.} 
}

\affiliations{
    \textsuperscript{1}Shenzhen International Graduate School, Tsinghua University \\
    \textsuperscript{2}Tencent, Shenzhen, China \\
    \textsuperscript{3}School of Artificial Intelligence, Shanghai Jiao Tong University \\
    \textsuperscript{4}Department of Computer Science, National University of Singapore\\
    xiaox@sz.tsinghua.edu.cn, peilinzhao@sjtu.edu.cn
}
\makeatletter
\def\@copyrightspace{\relax}
\makeatother

\begin{document}

\maketitle

\begin{abstract}
We present \textbf{CrystalDiT}, a diffusion transformer for crystal structure generation that achieves state-of-the-art performance 
by challenging the trend of architectural complexity. 
Instead of intricate, multi-stream designs, CrystalDiT employs a unified transformer that imposes a powerful inductive bias: treating lattice and atomic properties as a single, interdependent system. 
Combined with a periodic table-based atomic representation and a balanced training strategy, our approach achieves 8.78\% SUN (Stable, Unique, Novel) rate on MP-20, substantially outperforming recent methods including FlowMM (4.21\%) and MatterGen (3.66\%). Notably, CrystalDiT generates 63.28\% unique and novel structures while maintaining comparable stability rates, demonstrating that architectural simplicity can be more effective than complexity for materials discovery. Our results suggest that in data-limited scientific domains, carefully designed simple architectures outperform sophisticated alternatives that are prone to overfitting.
\end{abstract}
\begin{links}
    \link{Code}{https://github.com/hanyi2021/CrystalDiT.git}
\end{links}
\section{Introduction}

Materials discovery limits technological advancement in energy and sustainability \cite{jain2013commentary,merchant2023scaling}. Traditional screening approaches constrain exploration to known structures \cite{curtarolo2012aflow}.

Generative models offer a transformative alternative by directly proposing novel crystal structures. Recent methods have achieved promising results through various sophisticated approaches: ADiT employs two-stage latent diffusion with separate VAE and DiT components \cite{joshi2025all}, FlowMM uses Riemannian flow matching to handle crystal symmetries \cite{miller2024flowmm}, and MatterGen introduces joint diffusion processes with equivariant score networks and adapter modules \cite{zeni2025generative}. While these approaches demonstrate the potential of generative modeling, their architectural complexity raises questions about necessity and effectiveness.

While transformer architectures have found success across diverse domains 
\cite{feng2024latent,yi2023etdock,gong2024scaling,chen2024sdformer,wu2020adversarial,fang2024spatio,chen2025msdformer,shibo2025hdt,wang2025ntpp,xu2021relation}, 
the design principles for data-limited scientific applications remain underexplored.

We identify two critical issues with current approaches. First, evaluation metrics emphasize structural similarity to training data, inadvertently penalizing the novelty essential for discovery \cite{xie2021crystal}. Models optimized for traditional validity metrics excel at reproducing known patterns but struggle with exploration. Second, the trend toward architectural complexity may be counterproductive in materials science, where datasets are small and biased toward known stable structures. Unlike vision or language tasks with massive datasets, complex models risk overfitting rather than enabling generalization.

We present CrystalDiT, exploring whether simplified architectures can outperform complex alternatives in crystal generation. To rigorously test this hypothesis, we develop both a simple unified architecture and a complex dual-stream variant for direct comparison. Our key contributions are:

\textbf{Simplified Architecture}: A unified diffusion transformer processes all crystal information through joint attention, contrasting with our dual-stream alternative that uses separate processing pathways with cross-attention mechanisms.

\textbf{Chemical Representation}: A two-dimensional atomic encoding using periodic table positions (period, group) naturally captures chemical relationships without architectural complexity.

\textbf{Balanced Evaluation}: A composite score explicitly optimizes the trade-off between generation quality and discovery potential, addressing the evaluation-objective mismatch.

Our findings suggest that in data-limited scientific domains, carefully designed simple architectures with domain-specific representations outperform complex alternatives, challenging assumptions about the necessity of architectural sophistication.

\section{Related Work}

Crystal generation has become a central focus in the field of materials informatics. To address the limitations of traditional computational materials discovery methods, such as their effectiveness and high computational cost~\cite{pickard2011ab,curtarolo2012aflow,yamashita2018crystal,wang2021predicting}, recent approaches have increasingly leveraged deep learning-based generative models.

CDVAE~\cite{xie2021crystal} was among the first to predict three key components (the number and type of atoms and the lattice structure) as an initial approximation of a material's structure, which is subsequently refined using a diffusion-based approach~\cite{song2019generative,ho2020denoising}. DiffCSP~\cite{jiao2023diffcsp} is the first to employ a jointly equivariant diffusion paradigm (i.e., jointly diffusing the lattices and fractional coordinates) for crystal structure prediction (CSP). This method can be further extended to handle ab initio crystal generation by incorporating an additional discrete diffusion~\cite{austin2021structured} on atom types, with MatterGen~\cite{zeni2025generative} further refining this joint diffusion paradigm.

Diffusion Transformers (DiT)~\cite{peebles2023scalable} have shown remarkable capabilities in learning stable crystal structures through their powerful attention mechanisms and adaptive conditioning. However, this expressive power also introduces significant risks of overfitting in data-limited scientific domains, where models may memorize training patterns rather than learn generalizable principles for materials discovery. ADiT~\cite{joshi2025all} introduces a two-stage approach that first generates latent representations of crystal structures using an autoencoder, followed by applying the DiT architecture in the latent space for structure generation. 

Flow Matching techniques~\cite{lipman2022flow,lipman2024_fm_guide} have recently emerged as a highly effective alternative to diffusion models, primarily owing to their enhanced inference efficiency and the flexibility in defining prior distributions. In this context, FlowMM~\cite{miller2024flowmm} leverages joint Riemannian Flow Matching~\cite{chenriemannian} within Riemannian manifolds, providing improved handling of crystal periodicities. Subsequently, FlowLLM~\cite{sriram2024flowllm} extends this framework to large language models (LLMs), utilizing them as prior distributions for sampling the chemical formulas of meta-stable materials, with their corresponding structural configurations subsequently generated through Riemannian Flow Matching.

In another line of research, material symmetries have garnered significant attention in the context of crystal structure generation. Recently, DiffCSP++~\cite{jiao2024diffcsppp}, SymmCD~\cite{levy2025symmcd}, and WyFormer~\cite{kazeevwyckoff} have made notable advances by incorporating Wyckoff positions~\cite{wyckoff1922analytical} into the crystal generation task, enabling the generation of crystal structures with defined symmetries.
Despite these technical advances, existing approaches exhibit increasing architectural complexity with specialized multi-component designs. This motivates our investigation of whether unified, simplified architectures can achieve superior performance through more effective learning of lattice-atom relationships.

\section{Method}

We propose CrystalDiT, a simple yet effective diffusion transformer architecture for crystal structure generation. Our approach consists of four key components: (1) a novel two-dimensional atomic representation that captures chemical relationships through periodic table positioning, (2) a streamlined diffusion transformer architecture that processes crystal structures through unified attention mechanisms, (3) a balanced model selection strategy that replaces traditional validation, and (4) a probabilistic atomic decoding procedure for inference. This section details each component of our methodology.

\subsection{Crystal Representation}

Effective crystal structure representation is crucial for training diffusion models that can generate stable and novel materials. Traditional approaches either use graph-based representations that scale quadratically with the number of atoms, or rely on atomic number encodings that ignore chemical relationships. While recent work like UniMat \cite{yang2023scalable} proposed periodic table-based representations, their 4D tensor approach requires pre-allocating space for every possible element in the periodic table, leading to sparse representations where most positions remain unoccupied. Moreover, their method necessitates defining maximum atom counts per element type and specialized tensor operations across chemical dimensions. In contrast, our approach provides a more compact representation that only encodes 20 atoms and enables unified processing through standard transformer operations.

We introduce a simplified yet more chemically meaningful crystal representation that builds upon periodic table structure while incorporating domain-specific insights from materials science.

\subsubsection{Two-Dimensional Atomic Representation}

Instead of representing atoms by their atomic numbers, we encode each atom using its position in the periodic table: the period (row) and group (column). This encoding is motivated by the fundamental principle that elements in the same period share similar electron shell configurations, while elements in the same group exhibit similar chemical properties.

Specifically, for an atom with atomic number $Z$, we map it to a tuple $(r, c)$ where $r \in [0, 7]$ represents the period and $c \in [0, 18]$ represents the group. Here, $r = 0$ and $c = 0$ correspond to a special "null atom" representing empty positions in crystals with fewer than 20 atoms. For valid elements, $r \in [1, 7]$ and $c \in [1, 18]$, with lanthanides and actinides using fractional group numbers to preserve their unique positions. We then normalize these values to $[-1, 1]$:

\begin{align}
r_{\text{norm}} &= \frac{2r}{7} - 1 \\
c_{\text{norm}} &= \frac{2c}{18} - 1
\end{align}

This representation offers several advantages: (1) it naturally captures chemical similarity through spatial proximity in the periodic table, (2) it provides a continuous embedding space that facilitates diffusion modeling, (3) it reduces dimensionality while preserving chemical meaning, and (4) it seamlessly handles variable-sized crystal structures through the null atom representation at the origin $(0, 0)$.

\subsubsection{Normalized Lattice Parameterization}

For lattice vectors, we adopt a normalization strategy that addresses the wide range of lattice parameter values in real materials. Given the lattice matrix $\bm{L} \in \mathbb{R}^{3 \times 3}$, we normalize by the maximum length scale observed in the MP-20 dataset:

\begin{equation}
\bm{L}_{\text{norm}} = \frac{\bm{L}}{L_{\max}}
\end{equation}

where $L_{\max} = 46.7425$ \AA~from our analysis of the MP-20 dataset \cite{xie2021crystal}, which contains 45,231 metastable crystal structures with up to 20 atoms spanning 89 element types.

\subsubsection{Complete Structure Representation}

Our complete crystal representation combines the normalized lattice vectors and atomic features. For a crystal with $N$ atoms ($N \leq 20$ for MP-20):

\begin{align}
\bm{L}_{\text{norm}} &\in \mathbb{R}^{3 \times 3} \quad \text{(normalized lattice)} \\
\bm{A} &\in \mathbb{R}^{20 \times 5} \quad \text{(atomic features)}
\end{align}

where each row of $\bm{A}$ contains $[r_{\text{norm}}, c_{\text{norm}}, x, y, z]$ representing the normalized period, normalized group, and fractional coordinates. For crystals with fewer than 20 atoms, we pad with ``null" atoms using $[-1, -1, -1, -1, -1]$.

\subsection{Architecture Design}

Our CrystalDiT architecture is designed around the principle that simplicity leads to better generalization in crystal generation tasks. Unlike complex multi-stream architectures, we employ a unified approach that processes all crystal information through a single, streamlined transformer pathway.

\subsubsection{Unified DiT Architecture}

The model consists of three main components: (1) Crystal structure embedding that maps lattice vectors and atomic features into a shared hidden space, (2) A sequence of 18 DiT blocks that process the combined representation through unified self-attention, and (3) Specialized output heads that generate noise predictions for both atomic and lattice components.

The key insight is that by processing atomic and lattice features together in a single attention pathway, the model can naturally learn the complex interdependencies between atomic positions and lattice parameters without requiring explicit cross-attention mechanisms. This unified approach enforces a strong inductive bias: treating lattice and atomic properties as a single, interdependent system, which aligns with the physical reality that crystal properties emerge from the interplay between atomic composition and lattice geometry.

Crystal structures are embedded into a hidden space of dimension $d = 512$. The lattice vectors and atomic features are processed through separate linear embedding layers with positional and type encodings. The embedded features are concatenated to form a combined representation $\bm{H}_{\text{combined}} \in \mathbb{R}^{23 \times d}$ (20 atoms + 3 lattice vectors).

This combined representation is processed through $L = 18$ identical DiT blocks, each incorporating time-conditional adaptive layer normalization (AdaLN) to modulate features based on the diffusion time step. Finally, specialized output heads generate noise predictions for atomic features (5D) and lattice vectors (3D per vector).

\begin{figure*}[tb]
\centering
\includegraphics[width=1\textwidth]{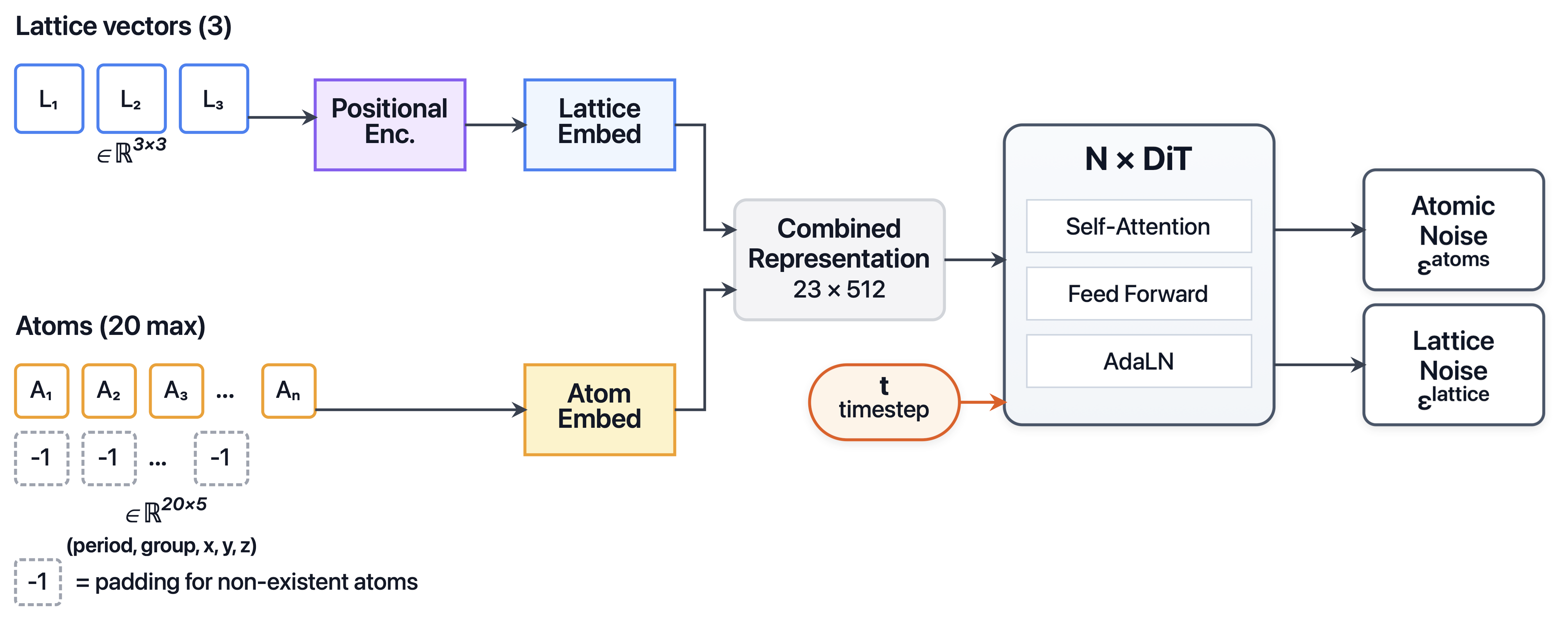}
\caption{CrystalDiT unified architecture. Input crystal structures are embedded into a combined 23-token sequence (3 lattice vectors + 20 atoms), processed through N DiT blocks with unified self-attention, and decoded to atomic and lattice noise predictions. The architecture treats all crystal components as a single interdependent system.}
\label{fig:architecture}
\end{figure*}

\subsubsection{Architecture Comparison}

In contrast to our unified approach shown in Figure~\ref{fig:architecture}, we also implement a complex dual-stream variant for direct comparison. This architecture employs cascaded processing: 12 atom-only DiT blocks process atomic features independently, followed by 2 lattice-only blocks for lattice vectors, and finally 2 joint blocks with bidirectional cross-attention mechanisms to fuse information between streams. Unlike our simple unified approach that processes all features together from the start, this cascaded design separates atomic and lattice processing pathways with specialized modules. However, this architectural complexity leads to overfitting in the data-limited crystal generation domain, as evidenced by lower unique and novel generation rates despite achieving higher individual validity metrics.

\subsection{Training Objective and Model Selection}

Our training approach combines standard diffusion objectives with a novel model selection strategy that addresses the unique challenges of crystal generation.

\subsubsection{Diffusion Loss Function}

We employ a Gaussian diffusion process with $T = 1000$ time steps and a linear noise schedule. Our model learns to predict the noise $\bm{\epsilon}$ added at each time step $t$ using a weighted loss function:

\begin{align}
\mathcal{L} = \mathcal{L}_{\text{lattice}} + \lambda \cdot \mathcal{L}_{\text{atoms}}
\end{align}

where:
\begin{align}
\mathcal{L}_{\text{lattice}} &= \mathbb{E}_{t,\bm{\epsilon}}\left[\|\bm{\epsilon}_{\text{lattice}} - \bm{\epsilon}_{\theta}^{\text{lattice}}(\bm{L}_t, \bm{A}_t, t)\|^2\right] \\
\mathcal{L}_{\text{atoms}} &= \mathbb{E}_{t,\bm{\epsilon}}\left[\bm{w}^T \odot \|\bm{\epsilon}_{\text{atoms}} - \bm{\epsilon}_{\theta}^{\text{atoms}}(\bm{L}_t, \bm{A}_t, t)\|^2\right]
\end{align}

We set $\lambda = 100$ to balance different scales, and use feature-specific weights $\bm{w} = [1.5, 2.0, 1.0, 1.0, 1.0]$ to emphasize period and group predictions.

\subsubsection{Balanced Model Selection Strategy}

Traditional checkpoint selection methods face fundamental limitations in crystal generation. Validation loss, while standard in machine learning, cannot capture the true stability and discovery potential of generated crystals, as it only measures reconstruction fidelity rather than physical plausibility. Alternative approaches employed by recent methods like ADiT \cite{joshi2025all} generate 1000 structures at each checkpoint and select models with the highest validity rates. However, this strategy inadvertently promotes overfitting to known stable patterns in the training data, leading to high validity scores but poor novelty rates - exactly opposite to what materials discovery requires.

Our \textbf{Balance Score} provides a principled alternative that explicitly optimizes the trade-off between generation quality and discovery potential during checkpoint selection:

\begin{equation}
\text{Balance Score} = \text{UN Rate} \times (\text{Quality Composite})^{\alpha}
\end{equation}

where $\alpha$ is a hyperparameter that controls the trade-off between generation quality and discovery potential. Higher $\alpha$ values emphasize quality for more reliable structures, while lower $\alpha$ values favor discovery of unique and novel crystals.

\textbf{UN Rate} measures the proportion of generated structures that are both unique and novel:
\begin{equation}
\text{UN Rate} = \frac{N_{\text{unique}} \cap N_{\text{novel}}}{N_{\text{total}}}
\end{equation}

\textbf{Quality Composite} is the geometric mean of four normalized quality metrics:
\begin{equation}
\text{Quality Composite} = (S_{\text{struct}} \times S_{\text{chem}} \times S_{\text{density}} \times S_{\text{elements}})^{1/4}
\end{equation}

Each component score is normalized to $[0,1]$ based on empirically observed ranges:
\begin{align}
S_{\text{struct}} &= \max\left(0, \min\left(1, \frac{V_{\text{struct}} - 0.95}{0.05}\right)\right) \\
S_{\text{chem}} &= \max\left(0, \min\left(1, \frac{V_{\text{chem}} - 0.8}{0.2}\right)\right) \\
S_{\text{density}} &= \max\left(0, \min\left(1, \frac{1.0 - D_{\text{density}}}{0.9}\right)\right) \\
S_{\text{elements}} &= \max\left(0, \min\left(1, \frac{1.0 - D_{\text{elements}}}{0.9}\right)\right)
\end{align}

We implement a multi-phase checkpoint selection strategy during training. We identify the models with the highest Balance Score from three distinct training phases: early (0-30\%), middle (31-60\%), and late (61-100\%) stages. This approach recognizes that optimal trade-offs between quality and discovery potential may emerge at different training stages, with early models potentially favoring exploration and later models emphasizing quality. The detailed training protocol is provided in Appendix C.

\subsection{Inference Procedure}

During inference, our model generates crystal structures through the standard DDPM sampling process, followed by a probabilistic atomic decoding procedure to convert continuous predictions to discrete atomic types.

\subsubsection{Probabilistic Atomic Decoding}

Our model predicts continuous values for atomic periods and groups, which must be mapped to discrete atomic numbers. We assign each candidate element a responsibility region and compute the probability using Gaussian integration.

For predicted continuous values $(r_{\text{pred}}, c_{\text{pred}})$ and candidate element $z$ at normalized position $(r_z, c_z)$, the mapping probability is:

\begin{align}
P(z|r_{\text{pred}}, c_{\text{pred}}) &= P_r(r_{\text{pred}}|r_z) \times P_c(c_{\text{pred}}|c_z) \\
P_r(r_{\text{pred}}|r_z) &= \int_{r_{\text{lower}}}^{r_{\text{upper}}} \frac{1}{\sqrt{2\pi\sigma^2}} \exp\left(-\frac{(x - r_{\text{pred}})^2}{2\sigma^2}\right) dx \\
P_c(c_{\text{pred}}|c_z) &= \int_{c_{\text{lower}}}^{c_{\text{upper}}} \frac{1}{\sqrt{2\pi\sigma^2}} \exp\left(-\frac{(x - c_{\text{pred}})^2}{2\sigma^2}\right) dx
\end{align}

where $\sigma = 0.1$ controls the mapping sharpness. The integration bounds define each element's responsibility region:

\begin{align}
r_{\text{upper}} &= \begin{cases}
r_z + \Delta_r & \text{if } r_z < 1 \\
+\infty & \text{if } r_z = 1
\end{cases} \\
r_{\text{lower}} &= \begin{cases}
-\infty & \text{if } r_z = -1 \\
r_z - \Delta_r & \text{if } r_z > -1
\end{cases}
\end{align}

with similar bounds for groups, where $\Delta_r = 1/7$ and $\Delta_c = 1/18$ are the discretization intervals. The final atomic number is selected as:
$$z^* = \arg\max_z P(z|r_{\text{pred}}, c_{\text{pred}})$$

For the null atom at $(-1, -1)$, its region extends to $(-\infty, -1+\Delta_r] \times (-\infty, -1+\Delta_c]$, allowing natural handling of empty positions.

\begin{algorithm}[tb]
\caption{CrystalDiT Generation with Probabilistic Atomic Decoding}
\label{alg:crystaldit_generation}
\small
\begin{algorithmic}[1] 
\STATE \textbf{Input:} Timesteps $T$, batch size $B$, model $\theta$
\STATE \textbf{Output:} Crystal structures $\{\text{Structure}_i\}_{i=1}^B$
\STATE 
\STATE Sample initial noise: $\bm{Z}^{(0)} \sim \mathcal{N}(\mathbf{0}, \mathbf{I})$ 
\FOR{$t = T, T-1, \ldots, 1$}
    \STATE $\bm{\epsilon} = \text{CrystalDiT}_\theta(\bm{Z}^{(t)}, t)$ 
    \STATE $\bm{Z}^{(t-1)} = \text{DDPM\_step}(\bm{Z}^{(t)}, \bm{\epsilon}, t)$ 
\ENDFOR
\STATE Extract lattice: $\bm{L} = \bm{Z}^{(0)}_{[:,:3,:]} \times L_{\max}$
\STATE Extract atoms: $\bm{A} = \bm{Z}^{(0)}_{[:,3:,:]}$
\FOR{each sample $i = 1, \ldots, B$}
    \STATE Initialize: $atomic\_numbers_i = [\,]$, $coords_i = [\,]$
    \FOR{each atom $j = 1, \ldots, 20$}
        \STATE $(r, c, x, y, z) = \bm{A}_{i,j,:5}$
        \STATE Compute $z^* = \arg\max_z P(z|r, c)$ using Eqs.~(16)-(20)
        \IF{$z^* > 0$}
            \STATE $atomic\_numbers_i$.append($z^*$)
            \STATE $coords_i$.append($[x, y, z] \bmod 1$)
        \ENDIF
        \COMMENT{Skip if $z^* = 0$ (null atom)}
    \ENDFOR
    \STATE Construct Structure$_i$ from $atomic\_numbers_i$ and $coords_i$
\ENDFOR
\end{algorithmic}
\end{algorithm}

\section{Experiments and Results}

We conduct comprehensive experiments to evaluate our CrystalDiT approach against state-of-the-art crystal generation methods. Our evaluation uses established metrics for cross-method comparison, while our Balance Score is specifically employed for checkpoint selection during our model training to optimize the trade-off between generation quality and discovery potential.

\subsection{Experimental Setup}
\textbf{Unified Evaluation Protocol:} To ensure fair comparison, we re-evaluate all baseline methods using a unified testing protocol with identical evaluation metrics, DFT calculation parameters, and statistical sampling procedures. This approach eliminates potential discrepancies arising from different evaluation implementations across original papers.

\subsubsection{Dataset and Preprocessing}
We use the MP-20 dataset \cite{xie2021crystal}, which contains 45,231 metastable crystal structures from the Materials Project with up to 20 atoms spanning 89 element types. Following standard practice, we use the established train/test split and preprocess structures using our two-dimensional atomic representation and lattice normalization.

\subsubsection{Model Configurations}
We evaluate three model variants:

\textbf{CrystalDiT (Simple):} Our main model with unified attention processing. Architecture: $d=512$, $L=18$ layers, 8 attention heads. Model size: 330MB.

\textbf{CrystalDiT (Complex):} A dual-stream architecture for comparison, featuring separate atom and lattice processing streams with cross-attention mechanisms. This architecture uses cascaded processing: 12 atom-only DiT blocks → 2 lattice-only DiT blocks → 2 joint DiT blocks with bidirectional cross-attention. Model size: 370MB (parameter count controlled to be similar to the simple version).

Both models are trained for 50,000 epochs with batch size 256, learning rate $1 \times 10^{-4}$ using 8 V100 GPUs over 4 days.

\subsubsection{Baseline Methods}
We compare against five state-of-the-art methods: DiffCSP \cite{jiao2023diffcsp}, FlowMM \cite{miller2024flowmm}, DiffCSP++ \cite{jiao2024diffcsppp}, and MatterGen \cite{zeni2025generative} using pretrained checkpoints or official implementations; ADiT \cite{joshi2025all} using 10,000 pre-generated structures from their official repository.

\subsubsection{Evaluation Protocol}
Following the evaluation framework established by FlowMM \cite{miller2024flowmm}, we generate 10,000 structures from each method and compute comprehensive metrics:

\textbf{Validity Metrics:}
\begin{itemize}
\item \textit{Structural validity}: Percentage of crystals with all pairwise atomic distances $\geq 0.5$ \AA~and crystal volume $\geq 0.1$ \AA$^3$
\item \textit{Compositional validity}: Percentage satisfying charge neutrality and electronegativity balance via SMACT \cite{davies2019smact}, using oxidation state enumeration and Pauling electronegativity rules
\end{itemize}

\textbf{Distribution Metrics:}
\begin{itemize}
\item \textit{Density distance} ($d_{\rho}$): Wasserstein distance between generated and test set density distributions
\item \textit{Elements distance} ($d_{\text{elem}}$): Wasserstein distance between generated and test set element occurrence frequency distributions
\end{itemize}

\textbf{Discovery Metrics:}
\begin{itemize}
\item \textit{Uniqueness}: Structures deemed distinct by PyMatGen's StructureMatcher \cite{ong2013python}
\item \textit{Novelty}: Structures not matching any MP-20 training set crystal via StructureMatcher
\item \textit{UN Rate}: Fraction of structures that are simultaneously unique and novel
\end{itemize}

\textbf{Stability Assessment:} We adapt the protocol established by FlowMM \cite{miller2024flowmm} for DFT evaluation. Due to computational resource limitations, we randomly sample 500 UN structures for stability assessment (compared to FlowMM's evaluation of all structures). To quantify sampling uncertainty, we repeat this sampling three times independently for CrystalDiT, FlowMM, and MatterGen, reporting mean$\pm$std across samples. Each sampled structure undergoes:

\begin{enumerate}
\item Pre-relaxation using CHGNet \cite{deng2023chgnet} ML potential
\item DFT relaxation using VASP with MPRelaxSet parameters \cite{jain2013commentary}
\item Energy above hull calculation against Matbench Discovery convex hull \cite{riebesell2023matbench}
\item Classification: Stable ($E_{\text{hull}} < 0.0$ eV/atom), Metastable ($E_{\text{hull}} < 0.1$ eV/atom)
\end{enumerate}

\textbf{Final Discovery Metrics:}
\begin{itemize}
\item \textit{SUN Rate}: UN Rate × Stable Rate among UN structures
\item \textit{MSUN Rate}: UN Rate × Metastable Rate among UN structures
\end{itemize}

Detailed evaluation parameters and implementation specifics are provided in Appendix A.

\subsection{Main Results}

Table~\ref{tab:main_results} presents our comprehensive comparison against state-of-the-art methods. The results reveal several key insights about the effectiveness of different approaches for crystal generation.
\begin{table*}[t]
\centering
\resizebox{\textwidth}{!}{ 
\begin{tabular}{l|cc|cc|c|cc|cc}
\toprule
\textbf{Method} & \textbf{Struct.} & \textbf{Chem.} & \textbf{$d_{\rho}$} & \textbf{$d_{\text{elem}}$} & \textbf{UN Rate} & \textbf{Stable} & \textbf{Metastable} & \textbf{SUN} & \textbf{MSUN} \\
 & \textbf{Valid (\%)} & \textbf{Valid (\%)} & \textbf{$\downarrow$} & \textbf{$\downarrow$} & \textbf{(\%)} & \textbf{in UN (\%)} & \textbf{in UN (\%)} & \textbf{(\%)} & \textbf{(\%)} \\
\midrule
MP-20(train)$^*$ & 100.0 & 90.55 & 0.214 & 0.049 & - & 44.07$^*$ & 100.0$^*$ & - & - \\
\midrule
DiffCSP & 99.90 & 82.52 & 0.347 & 0.369 & 87.17 & 4.00 & 23.80 & 3.49 & 20.75 \\
FlowMM & 99.22 & 82.09 & 0.185 & 0.128 & 87.66 & 4.80$_{\pm0.20}$ & 23.69$_{\pm0.10}$ & 4.21$_{\pm0.18}$ & 20.77$_{\pm0.09}$ \\
DiffCSP++ & 99.96 & 84.74 & \textbf{0.135} & 0.453 & 87.62 & 3.80 & 21.80 & 3.33 & 19.10 \\
MatterGen & \textbf{99.99} & 83.62 & 0.393 & 0.207 & \textbf{89.89} & 4.07$_{\pm0.32}$ & 26.90$_{\pm0.71}$ & 3.66$_{\pm0.28}$ & 24.18$_{\pm0.63}$ \\
ADiT & 99.58 & \textbf{90.83} & 0.179 & \textbf{0.082} & 37.08 & 7.40 & 36.40 & 2.74 & 13.50 \\
\midrule
CrystalDiT (Complex) & 98.39 & 89.44 & 0.271 & 0.115 & 40.28 & \textbf{15.80} & \textbf{55.20} & 6.36 & 22.24 \\
\textbf{CrystalDiT (Simple)} & 97.79 & 87.02 & 0.459 & 0.211 & 63.28 & 13.87$_{\pm1.21}$ & 40.93$_{\pm1.51}$ & \textbf{8.78}$_{\pm\textbf{0.74}}$ & \textbf{25.90}$_{\pm\textbf{0.95}}$ \\
CrystalDiT (1D atomic) & 97.34 & 86.82 & 0.499 & 0.276 & 78.47 & 8.00 & 31.00 & 6.28 & 24.33 \\
\bottomrule
\end{tabular}
} 
\caption{Comprehensive comparison on MP-20. Best results in bold. For methods with bootstrap sampling (indicated by $\pm$), we report mean$\pm$std over 3 independent samples of 500 UN structures each for DFT evaluation. $^*$Rates for MP-20(train) represent proportions across all training structures.}
\label{tab:main_results}
\end{table*}
\textbf{Note on Baseline Results:} Our re-evaluation using unified protocols yields results that differ from some originally reported values in the literature. Detailed analysis of these differences is provided in Appendix A.

Our simple CrystalDiT achieves the highest SUN rate (8.78\%) and MSUN rate (25.90\%), substantially outperforming recent methods including FlowMM (4.21\%) and MatterGen (3.66\%). Notably, CrystalDiT generates 63.28\% unique and novel structures while maintaining comparable stability rates, demonstrating that architectural simplicity can be more effective than complexity for materials discovery. Different methods exhibit distinct trade-offs: ADiT achieves high validity scores but suffers from poor novelty (37.08\% UN rate) due to overfitting, while FlowMM and MatterGen generate novel structures with lower stability rates.

\subsection{Architecture Comparison Analysis}

The comparison between our simple and complex architectures reveals fundamental insights about generative modeling for scientific applications. Despite sophisticated cross-attention mechanisms, the complex dual-stream architecture underperforms the simple version across discovery metrics (6.36\% vs 8.78\% SUN rate). The implementation and analysis of the dual-stream architecture are provided in Appendix B.

This result challenges prevailing assumptions about architectural sophistication in machine learning. The complex model achieves better individual quality metrics but significantly lower UN rates, indicating that architectural complexity promotes overfitting in crystal generation tasks. The simple unified attention mechanism appears more effective at learning generalizable patterns rather than memorizing training data distributions.

Similarly, ADiT's two-stage approach (autoencoder + latent DiT) achieves excellent validity scores but poor novelty (37.08\% UN rate), demonstrating that architectural sophistication without careful consideration of the discovery objective can be counterproductive in materials science applications.

\subsection{Training Dynamics and Model Selection}

Table~\ref{tab:simple_training} shows the evolution of our simple CrystalDiT model across different training epochs, demonstrating the importance of our balance score for checkpoint selection during training. Note that our Balance Score is used exclusively for selecting the best checkpoint during training of our CrystalDiT models, not for comparing different methods. All cross-method comparisons use the standard SUN/MSUN discovery metrics.

\begin{table}[htbp]
\centering
\small
\begin{tabular}{c|cc|cc|c}
\toprule
\textbf{Epoch} & \textbf{Struct.} & \textbf{Chem.} & \textbf{$d_{\rho}$} & \textbf{$d_{\text{elem}}$} & \textbf{UN} \\
 & \textbf{Valid} & \textbf{Valid} & & & \textbf{Rate} \\
 & \textbf{(\%)} & \textbf{(\%)} & & & \textbf{(\%)} \\
\midrule
10k & 96.77 & 83.09 & 0.551 & 0.178 & 80.72 \\
20k & 97.51 & 87.07 & 0.170 & 0.249 & 73.31 \\
30k & 97.99 & 86.74 & 0.308 & 0.231 & 63.30 \\
40k & 98.44 & 89.33 & 0.592 & 0.208 & 62.32 \\
50k & 98.28 & 89.40 & 0.166 & 0.266 & 57.37 \\
\bottomrule
\end{tabular}
\caption{Training progression of CrystalDiT (Simple) showing deteriorating UN rate despite improving validity metrics.}
\label{tab:simple_training}
\end{table}

The simple model shows a clear pattern: as training progresses, validity metrics improve, but the UN rate steadily decreases from 80.72\% to 57.37\%. Traditional validation methods focusing on validity would select the final checkpoint, but our balance score correctly identifies earlier checkpoints with better discovery potential. This demonstrates the critical importance of balanced evaluation for materials discovery applications.

\subsection{Component Analysis}
We evaluate the contribution of our two-dimensional atomic representation by comparing against traditional one-dimensional atomic number encoding on the simple CrystalDiT architecture. Detailed implementation of the 1D atomic representation is provided in Appendix B.

As shown in Table~\ref{tab:main_results}, our two-dimensional periodic table-based representation (CrystalDiT Simple: 8.78\% SUN rate) significantly outperforms the one-dimensional encoding (CrystalDiT 1D atomic: 6.28\% SUN rate). While the one-dimensional approach achieves higher UN rate (78.47\% vs 63.28\%), the generated structures exhibit lower stability rates (8.00\% vs 13.87\% stable rate), resulting in inferior final discovery performance. This demonstrates that chemical knowledge embedded in the periodic table structure enhances the quality of generated crystals for materials discovery.

We additionally conduct ablation studies on architecture depth, finding that 18 layers provide the optimal balance between model capacity and generalization. Detailed results are provided in Appendix B.

\subsection{Energy Distribution Analysis}

\begin{figure}[t]
\centering
\includegraphics[width=1.0\columnwidth]{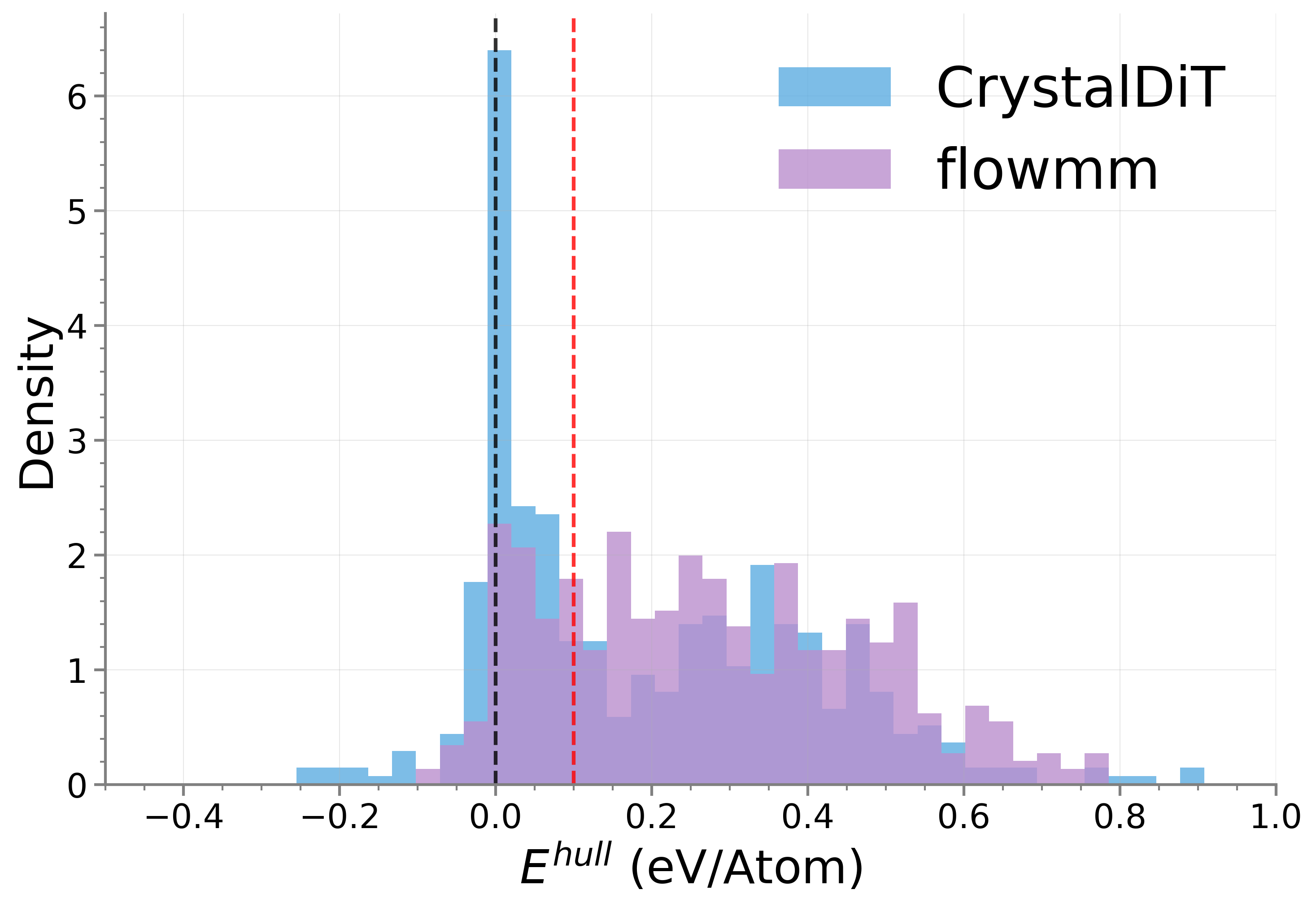}
\caption{Energy distribution comparison. Black and red dashed lines mark stability ($E^{\text{hull}} = 0$) and metastability ($E^{\text{hull}} = 0.1$ eV/atom) thresholds. CrystalDiT generates more stable and metastable structures. Full comparisons in Appendix C.}
\label{fig:energy_distributions}
\end{figure}

Figure~\ref{fig:energy_distributions} presents energy distribution comparison between CrystalDiT and FlowMM as a representative baseline method. Our analysis reveals that CrystalDiT demonstrates superior ability to generate thermodynamically favorable structures compared to FlowMM. CrystalDiT shows a pronounced peak in the stable region ($E^{\text{hull}} < 0$), indicating that our simplified architecture effectively learns to generate energetically favorable crystal structures. The energy distribution provides crucial validation that CrystalDiT not only generates more unique and novel structures but also ensures these structures are more likely to be thermodynamically viable for practical materials applications.

Similar patterns are observed when comparing against other baseline methods (ADiT, MatterGen, DiffCSP++), with CrystalDiT consistently producing more stable and metastable structures. These comprehensive energy distribution analyses are provided in Appendix C, reinforcing our main findings that simple, well-designed architectures with appropriate domain knowledge can outperform complex alternatives in crystal generation tasks.

\subsection{Scaling to Larger Structures}
CrystalDiT achieves 6.73\% SUN rate on MPTS-52 (up to 52 atoms), demonstrating effective generalization with only 2\% degradation compared to MP-20. Detailed results are in Appendix C.

\section{Conclusion}

We presented CrystalDiT, demonstrating that simplified architectures 
can substantially outperform complex alternatives for crystal generation. 
Our unified diffusion transformer achieves an 8.78\% SUN rate on MP-20, 
outperforming existing methods through three key contributions: 
(1) a simplified architecture using unified attention mechanisms; 
(2) a two-dimensional atomic representation using periodic table positions; 
and (3) a balanced model selection strategy optimizing generation quality 
and discovery potential.

Beyond generating isolated structures, important directions for future 
research include extending our framework to material-molecule interaction 
systems \cite{bian2026affinity,bian2026deep}, which could enable discovery 
of functional materials for catalysis, drug delivery, and energy applications. 
Additionally, incorporating target property constraints into the generation 
process would allow direct design of materials with desired characteristics, 
further advancing AI-driven materials discovery.

\section*{Acknowledgments}

This work was supported by the Natural Science Foundation of Guangdong Province (grant no. 2025A1515011946) and the National University of Singapore School of Computing (grant no. A-0010308-00-00 for YB). Part of this work was conducted when authors Xiaohan Yi and Guikun Xu were at Tencent AI Lab. We acknowledge computational resources from Tencent, thank Tao Chen for insightful discussions, and the anonymous reviewers for their constructive feedback.
\bibliography{aaai2026}

\onecolumn
\appendix

\setcounter{secnumdepth}{2}

\makeatletter
\renewcommand{\section}{\@startsection{section}{1}{\z@}%
  {-3.5ex \@plus -1ex \@minus -.2ex}%
  {2.3ex \@plus.2ex}%
  {\normalfont\Large\bfseries}}  
\renewcommand{\subsection}{\@startsection{subsection}{2}{\z@}%
  {-3.25ex\@plus -1ex \@minus -.2ex}%
  {1.5ex \@plus .2ex}%
  {\normalfont\large\bfseries}}  
\makeatother

\section{Evaluation Metrics}

\subsection{Evaluation Metrics Explanation}

Our comprehensive evaluation framework generates 10,000 crystal structures for each method and consists of two main stages: (1) structural and compositional assessment, and (2) thermodynamic stability evaluation. This section provides detailed implementation specifications for reproducibility.

\subsubsection{Stage 1: Structural and Compositional Assessment}

\textbf{Initial Structure Parsing and Pre-filtering}

All 10,000 generated CIF files undergo initial parsing and systematic pre-filtering with specific validity criteria. The pre-filtering process applies the following sequential checks:

\begin{enumerate}
\item Parse each CIF file using PyMatGen's CifParser to extract structure information
\item Extract lattice parameters (lengths and angles) and atomic information
\item Apply pre-filtering criteria in the following order:
   \begin{itemize}
   \item \textbf{Atomic number range check}: All atomic numbers must be within [1, 104]
   \item \textbf{Positive lattice parameters}: All lattice lengths must be positive ($> 0$)
   \item \textbf{NaN/Inf value check}: No NaN or infinite values in lattice parameters, angles, or atomic coordinates
   \item \textbf{PyMatGen Structure construction}: Must successfully construct a valid Structure object
   \item \textbf{Minimum volume check}: Crystal volume must be $\geq 0.1$ \AA$^3$
   \end{itemize}
\item Structures failing any pre-filtering criterion are marked as invalid with specific error reasons
\item All percentage calculations use 10,000 as the denominator, treating pre-filtered structures as invalid
\end{enumerate}

The pre-filtering serves as a comprehensive quality gate to ensure that only physically meaningful crystal structures proceed to detailed evaluation. Structures with invalid atomic compositions, degenerate lattice parameters, numerical errors, or unrealistically small volumes are systematically excluded.

\textbf{Structural Validity Assessment}

Structural validity uses distance-based geometric constraints following the FlowMM protocol \cite{miller2024flowmm}:

\begin{align}
\text{Structural Validity} = \begin{cases}
\text{True} & \text{if } \min(\text{distance\_matrix}) \geq 0.5 \text{ \AA} \text{ and } V_{\text{cell}} \geq 0.1 \text{ \AA}^3 \\
\text{False} & \text{otherwise}
\end{cases}
\end{align}

where the distance matrix excludes self-interactions (diagonal elements set to infinity).

\textbf{Compositional Validity Assessment}

Chemical validity uses the SMACT framework  with timeout protection:

\begin{enumerate}
\item Extract elemental composition from Structure objects
\item Enumerate oxidation state combinations for all elements
\item Apply charge neutrality: $\sum_i n_i \cdot z_i = 0$
\item Apply Pauling electronegativity test
\item Metallic alloy exception: all-metallic structures are valid
\item 30-second timeout per structure to prevent infinite enumeration
\end{enumerate}

\textbf{Distribution Distance Metrics}

Property distribution fidelity is quantified using Wasserstein distances. From structures passing both structural and chemical validity tests, we randomly sample 1000 structures (or use all if fewer than 1000 are available) and compare against the MP-20 test set:

\begin{align}
d_{\rho} &= W_1(\rho_{\text{gen,sample}}, \rho_{\text{test}}) \\
d_{\text{elem}} &= W_1(E_{\text{gen,sample}}, E_{\text{test}})
\end{align}

where $\rho$ represents density distributions and $E$ represents element occurrence frequency distributions.

\textbf{Uniqueness and Novelty Assessment}

Both metrics use PyMatGen's StructureMatcher with default parameters:

\begin{itemize}
\item \texttt{ltol}: 0.2 (lattice parameter tolerance)
\item \texttt{stol}: 0.3 (site position tolerance)  
\item \texttt{angle\_tol}: 5° (lattice angle tolerance)
\item \texttt{primitive\_cell}: True, \texttt{scale}: True, \texttt{attempt\_supercell}: False
\end{itemize}

\textbf{Chemical System Grouping Optimization}: Structures are first grouped by chemical systems (sorted element types) before pairwise comparison. This optimization reduces computational complexity from $O(n^2)$ to $O(\sum_i n_i^2)$ without affecting results, as verified through testing.

\textbf{Single Element Filtering}: Structures containing only one element type are excluded from uniqueness and novelty calculations during the UN assessment stage. This filtering is applied specifically for UN rate calculations, not during the initial pre-processing stage.

\textbf{Uniqueness}: Structures are compared against all other generated structures within the same chemical system. A structure is unique if it shows no StructureMatcher match.

\textbf{Novelty}: Generated structures are compared against the MP-20 training set within the same chemical system, following FlowMM's chemical system optimization approach.

\begin{equation}
\text{UN Rate} = \frac{\text{Structures that are both unique and novel}}{10,000}
\end{equation}

\subsubsection{Stage 2: Thermodynamic Stability Evaluation}

\textbf{Candidate Selection and CHGNet Pre-relaxation}

From the UN structures identified in Stage 1, we randomly sample 500 structures for stability assessment. Each structure undergoes CHGNet  pre-relaxation with enhanced parameters:

\begin{itemize}
\item Maximum relaxation steps: 2,500 (vs. 1,000 in FlowMM \cite{miller2024flowmm})
\item Structure optimization: Both atomic positions and lattice parameters
\end{itemize}

\textbf{DFT Candidate Filtering}

Structures are selected for DFT calculations based on CHGNet-predicted energy above hull:
\begin{equation}
E_{\text{hull}}^{\text{CHGNet}} < 0.5 \text{ eV/atom}
\end{equation}

This expanded threshold (vs. FlowMM's 0.1 eV/atom) enables comprehensive stability assessment, selecting approximately 490 out of 500 UN structures for DFT calculations compared to FlowMM's ~261 structures.

\textbf{DFT Relaxation Protocol}

DFT calculations employ VASP \cite{kresse1996efficient} with Materials Project Relaxation Set (MPRelaxSet) parameters \cite{jain2013commentary}, including PBE-GGA exchange-correlation functional, PAW pseudopotentials, and appropriate k-point sampling and convergence criteria.

\textbf{Energy Above Hull Calculation}

Final stability assessment uses the latest Materials Project convex hull (December 2024 release) with Materials Project 2020 compatibility corrections:

\begin{align}
E_{\text{hull}} &= E_{\text{DFT}}^{\text{corrected}} - E_{\text{hull}}^{\text{MP}} \\
\text{Stable} &: E_{\text{hull}} < 0.0 \text{ eV/atom} \\
\text{Metastable} &: E_{\text{hull}} < 0.1 \text{ eV/atom}
\end{align}

\textbf{Final Discovery Metrics}

\begin{align}
\text{SUN Rate} &= \text{UN Rate} \times \frac{\text{Stable UN structures}}{500} \\
\text{MSUN Rate} &= \text{UN Rate} \times \frac{\text{Metastable UN structures}}{500}
\end{align}

\subsection{Results Analysis}

\subsubsection{MatterGen Evaluation Discrepancy Analysis}

Our MatterGen evaluation yields different results compared to the original paper due to dataset-specific evaluation protocols. Table \ref{tab:mattergen_comparison} compares novelty assessment using different reference datasets.

\begin{table}[h]
\centering
\caption{MatterGen Novelty Assessment: MP-20 vs. Alexandria-MP-20 Reference Datasets}
\label{tab:mattergen_comparison}
\begin{tabular}{l|cc}
\toprule
\textbf{Reference Dataset} & \textbf{MP-20 Train} & \textbf{Alexandria-MP-20 Train} \\
\midrule
Unique Rate (\%) & 96.38 & 96.38 \\
Novel Rate (\%) & 91.31 & 82.02 \\
UN Rate (\%) & 89.89 & 81.21 \\
\bottomrule
\end{tabular}
\end{table}

The discrepancy arises because MatterGen's original evaluation used the larger Alexandria-MP-20 dataset as the novelty reference. Alexandria-MP-20 contains 607,683 training structures compared to MP-20's 27,136 training structures, making novelty assessment more stringent. Since we evaluate MatterGen's performance on MP-20-trained models, comparison against the MP-20 training set provides the appropriate novelty assessment. Our results (89.89\% UN rate) represent the correct evaluation for MP-20-based model performance.

\subsubsection{FlowMM Evaluation Protocol Differences}

Our FlowMM re-evaluation yields higher stability rates than reported in the original paper due to two methodological improvements:

\textbf{Updated Convex Hull Database}: We employ the latest Materials Project convex hull (December 2024 release), which provides more accurate thermodynamic reference data compared to earlier versions used in the original FlowMM evaluation.

\textbf{Expanded DFT Candidate Selection}: Our protocol evaluates more structures for comprehensive stability assessment:

\begin{itemize}
\item FlowMM protocol: CHGNet $E_{\text{hull}} < 0.1$ eV/atom → 261/500 structures evaluated
\item Our protocol: CHGNet $E_{\text{hull}} < 0.5$ eV/atom → 490/500 structures evaluated
\end{itemize}

This expanded evaluation provides a more representative assessment of stability rates within the UN structure population, reducing sampling bias that could arise from overly restrictive pre-screening. The higher stability rates we observe reflect both improved thermodynamic references and more comprehensive population sampling.

\section{Ablation Study}

\subsection{Dual-Stream Architecture}

To validate our hypothesis that architectural simplicity outperforms complexity in crystal generation, we implement a sophisticated dual-stream variant featuring cascaded processing with cross-attention mechanisms. This complex architecture serves as a direct comparison to our unified approach.

\subsubsection{Architecture Design}

\begin{figure}[htbp]
\centering
\includegraphics[width=1.0\textwidth]{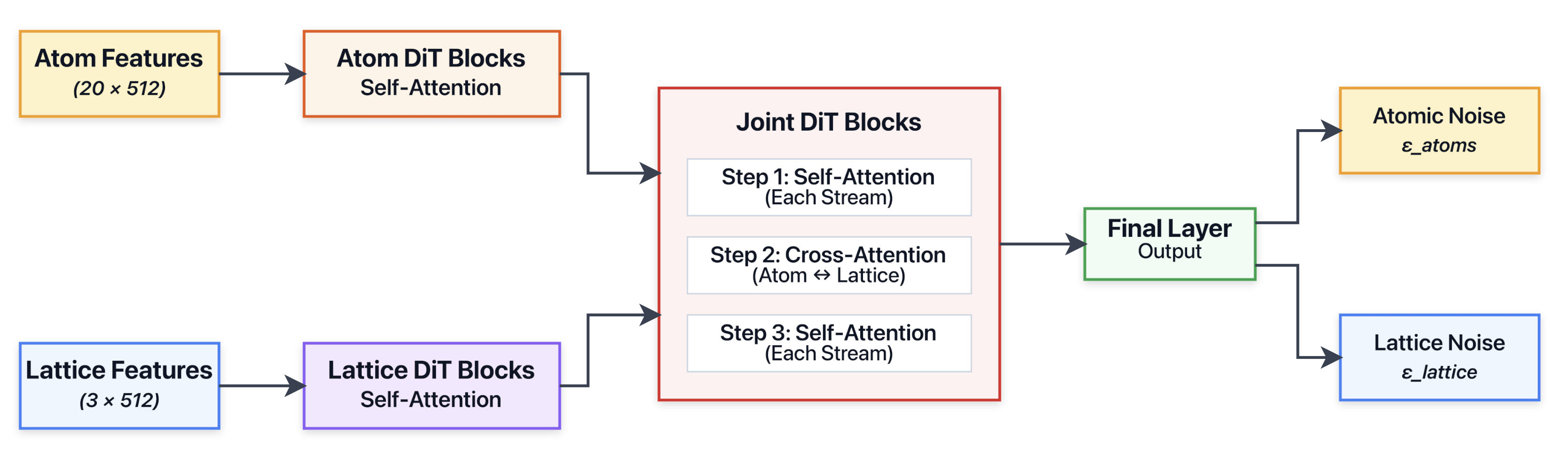}
\caption{Dual-stream cascaded architecture showing the three-stage processing: separate self-attention on 20 atomic tokens and 3 lattice tokens, followed by joint processing with sophisticated attention mechanisms.}
\label{fig:dual_stream_arch}
\end{figure}

The dual-stream architecture implements a sophisticated three-stage cascaded processing paradigm as illustrated in Figure~\ref{fig:dual_stream_arch}. After crystal embedding, the architecture processes 20 atomic tokens and 3 lattice vector tokens through separate dedicated pathways before sophisticated joint processing.

\textbf{Stage 1: Independent Atomic Processing.} The atomic pathway processes 20 atomic tokens through 12 dedicated DiT blocks using pure self-attention mechanisms. This allows the model to learn complex intra-atomic relationships and chemical patterns within the atomic ensemble without interference from lattice geometric constraints.

\textbf{Stage 2: Independent Lattice Processing.} In parallel, 3 lattice vector tokens are processed through 2 dedicated DiT blocks with self-attention. This separate pathway enables learning of lattice-specific geometric and crystallographic patterns independently from atomic composition information.

\textbf{Stage 3: Sophisticated Joint Processing.} The final stage employs 2 joint DiT blocks that implement a complex four-step attention mechanism within each block:

\begin{enumerate}
\item \textbf{Initial Self-Attention:} Separate self-attention for atomic and lattice features
\item \textbf{Bidirectional Cross-Attention:} Simultaneous lattice→atoms and atoms→lattice information exchange
\item \textbf{Post-Cross Self-Attention:} Additional self-attention after cross-modal information integration  
\item \textbf{Independent MLP Processing:} Separate feed-forward networks for both streams
\end{enumerate}

The attention flow within each joint block follows:
\begin{align}
\text{Atom}' &= \text{Atom} + \text{SelfAttn}(\text{Atom}) \\
\text{Lattice}' &= \text{Lattice} + \text{SelfAttn}(\text{Lattice}) \\
\text{Atom}'' &= \text{Atom}' + \text{CrossAttn}(\text{Atom}', \text{Lattice}') \\
\text{Lattice}'' &= \text{Lattice}' + \text{CrossAttn}(\text{Lattice}', \text{Atom}') \\
\text{Atom}_{out} &= \text{Atom}'' + \text{SelfAttn}(\text{Atom}'') + \text{MLP}(\text{Atom}'') \\
\text{Lattice}_{out} &= \text{Lattice}'' + \text{SelfAttn}(\text{Lattice}'') + \text{MLP}(\text{Lattice}'')
\end{align}

This cascaded design with sophisticated cross-attention mechanisms represents significant architectural complexity compared to our unified approach, requiring careful coordination between multiple processing streams and attention mechanisms.

\subsubsection{Computational Complexity Analysis}

We measured the actual computational overhead of both architectures under identical training conditions using 8× V100 GPUs with batch size 256.

\begin{table}[htbp]
\centering
\small
\begin{tabular}{lccc}
\toprule
\textbf{Architecture} & \textbf{Parameters} & \textbf{Time per Epoch} \\
\midrule
CrystalDiT (Simple) & 330MB & 5.39s  \\
CrystalDiT (Complex) & 370MB & 5.82s \\
\bottomrule
\end{tabular}
\caption{Empirical computational complexity comparison on 8× V100 GPUs with batch size 256.}
\label{tab:empirical_complexity}
\end{table}

The complex architecture introduces a modest 7.98\% training time overhead, primarily attributed to the bidirectional cross-attention computations in joint blocks. While this overhead appears manageable, the performance degradation is substantial: the complex model achieves 6.36\% SUN rate compared to 9.62\% for the simple architecture—a 34\% relative decrease in discovery performance.

This empirical analysis reveals a critical insight: the computational overhead of architectural complexity is relatively small, but the impact on generalization capability is severe. The cross-attention mechanisms, while computationally tractable, fundamentally alter the model's learning dynamics in ways that promote overfitting to training data patterns. This suggests that for scientific generative modeling tasks, the bottleneck lies not in computational resources but in the architectural inductive biases that guide model learning.

\subsubsection{Cross-Attention Mechanism Analysis}

The bidirectional cross-attention mechanism in joint blocks is designed to capture intricate relationships between atomic positions and lattice parameters. However, our analysis reveals that this sophisticated interaction modeling leads to overfitting behaviors. The complex model achieves higher individual validity scores (98.39\% structural, 89.44\% compositional) but significantly lower novelty rates, indicating that cross-attention mechanisms may promote memorization of training patterns rather than generalization to novel crystal structures.

\subsection{One-Dimensional Atomic Representation}

To evaluate the contribution of our two-dimensional periodic table-based encoding, we implement a baseline using traditional one-dimensional atomic number representation.

\subsubsection{Implementation Details}

The one-dimensional representation directly encodes atoms using their atomic numbers, normalized to the range [-1, 1]:

\begin{equation}
z_{\text{norm}} = \frac{2z}{Z_{\max}} - 1
\end{equation}

where $z$ is the atomic number and $Z_{\max} = 94$ represents the maximum atomic number in MP-20 dataset. Invalid atoms (empty positions) are represented as $z_{\text{norm}} = -1$.

The atomic feature vector becomes 4-dimensional: $[z_{\text{norm}}, x, y, z]$ instead of our 5-dimensional periodic table representation $[r_{\text{norm}}, c_{\text{norm}}, x, y, z]$. This encoding preserves atomic identity but loses the inherent chemical relationships embedded in the periodic table structure.

\subsubsection{Chemical Knowledge Integration}

The key difference between representations lies in chemical knowledge integration:

\textbf{1D Atomic Number:} Direct mapping from atomic numbers provides precise elemental identification but treats each element as independent. Chemical similarity between elements (e.g., alkali metals, transition metals) is not explicitly encoded in the representation space.

\textbf{2D Periodic Table:} Our approach embeds chemical relationships through spatial proximity in the normalized (period, group) space. Elements with similar properties naturally cluster in the embedding space, providing inductive bias for chemical reasoning.

\subsubsection{Performance Analysis}

The 1D representation achieves higher UN rates (78.47\% vs 63.28\%) but significantly lower stability rates (8.0\% vs 15.2\% stable rate among UN structures). This trade-off results in inferior final discovery metrics (6.2\% vs 9.62\% SUN rate), demonstrating that chemical knowledge embedded in periodic table structure enhances the practical utility of generated crystals.

\subsection{Architecture Depth Ablation}

To validate our choice of 18 DiT blocks, we conduct systematic ablation studies across different model depths. Table~\ref{tab:depth_ablation} presents comprehensive results.

\begin{table}[htbp]
\centering
\small
\begin{tabular}{c|ccc}
\toprule
\textbf{Depth} & \textbf{UN Rate (\%)} & \textbf{SUN (\%)} & \textbf{MSUN (\%)} \\
\midrule
6 layers   & 82.4 & 5.78 & 23.48 \\
12 layers  & 73.2 & 6.95 & 24.89 \\
18 layers  & 63.3 & 8.78 & 25.90 \\
24 layers  & 56.8 & 7.10 & 26.41 \\
\bottomrule
\end{tabular}
\caption{Architecture depth ablation study. 18 layers achieves optimal balance between exploration (UN rate) and stability (SUN rate).}
\label{tab:depth_ablation}
\end{table}

\subsubsection{Analysis}

The results reveal a clear trade-off between model capacity and generalization:

\textbf{Shallow Models (6 layers):} Achieve the highest UN rate (82.4\%) but lower SUN rate (5.78\%), indicating excessive exploration with insufficient quality control. The limited capacity fails to capture complex stability patterns.

\textbf{Optimal Depth (18 layers):} Provides the best discovery performance with 8.78\% SUN rate. This depth achieves optimal balance between generating novel structures and ensuring thermodynamic stability.

\textbf{Deep Models (24 layers):} Show overfitting behavior with decreased UN rate (56.8\%) and slightly reduced SUN rate (7.10\%) compared to 18 layers. The excessive capacity leads to memorization of training patterns, reducing exploration capability.

This systematic analysis validates our architectural choice and demonstrates that model depth is a critical hyperparameter in crystal generation tasks.

\subsection{Normalization Ablation}

To validate the importance of our lattice normalization strategy, we train CrystalDiT without the normalization described in Section 3.1.2.

\textbf{Results:} The unnormalized model achieves 49.4\% UN rate, 4.70\% SUN rate, and 18.04\% MSUN rate, substantially underperforming our normalized approach (63.3\% UN, 8.78\% SUN, 25.90\% MSUN).

\textbf{Analysis:} Without normalization, the wide range of lattice parameter values (from $\sim$3Å to 46Å in MP-20) creates training instabilities and prevents effective learning of lattice-atomic correlations. This validates that proper feature scaling is essential for diffusion models in materials science applications.

\subsection{Element Distribution Analysis}

\begin{figure}[htbp]
\centering
\includegraphics[width=0.8\textwidth]{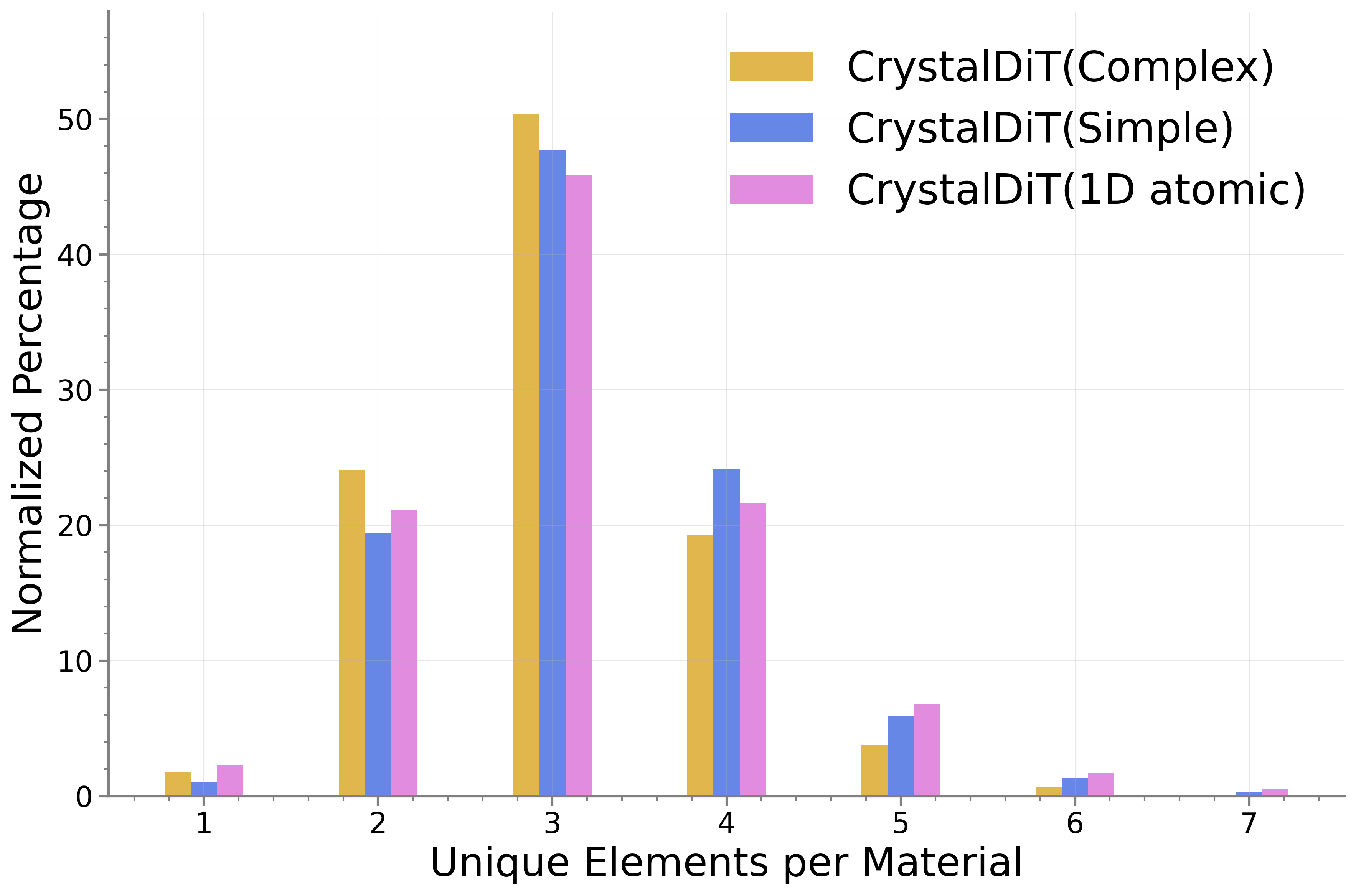}
\caption{Element count distribution comparison across CrystalDiT variants. The simple architecture generates more chemically diverse structures while maintaining stability.}
\label{fig:element_distribution}
\end{figure}

Figure~\ref{fig:element_distribution} presents the element count distribution across our three model variants, revealing important insights about chemical diversity and complexity handling.

\subsubsection{Distribution Characteristics}

All three variants show similar overall patterns with peak generation around 3-element compositions, consistent with the MP-20 training data distribution. However, subtle but important differences emerge:

\textbf{CrystalDiT (Complex):} Shows the most constrained distribution (mean: 3.016 elements, std: 0.851), with 50.38\% of structures containing exactly 3 elements. The tight distribution suggests the complex architecture tends toward conservative, training-data-like compositions.

\textbf{CrystalDiT (Simple):} Exhibits slightly higher diversity (mean: 3.199 elements, std: 0.925) with more structures containing 4+ elements (31.76\% vs 24.62\% for complex). The broader distribution indicates better exploration of chemical space while maintaining stability.

\textbf{CrystalDiT (1D atomic):} Demonstrates the highest variability (std: 1.016), but this increased diversity comes at the cost of stability, as evidenced by lower stable rates in our main results.

\subsubsection{Chemical Complexity vs. Stability Trade-off}

The element distribution analysis reveals a fundamental trade-off in crystal generation: while higher chemical diversity can lead to more novel discoveries, it must be balanced against thermodynamic stability. Our simple architecture with 2D periodic table representation achieves the optimal balance, generating moderately diverse chemical compositions that maintain high stability rates.

The complex architecture's constrained distribution aligns with its overfitting behavior—generating safer, training-data-like compositions but missing opportunities for novel discovery. Conversely, the 1D representation's excessive diversity without chemical guidance leads to unstable, impractical structures.

This analysis reinforces our main finding that architectural simplicity combined with appropriate chemical knowledge (2D periodic representation) provides the most effective approach for materials discovery applications.

\section{Additional Results}

\subsection{Balance Score Analysis and Training Dynamics}

To provide comprehensive validation of our Balance Score methodology and training approach, we present detailed analysis of training dynamics and checkpoint selection across different hyperparameter settings.

\subsubsection{Evaluation Protocol}

Our training evaluation follows a systematic protocol: we save checkpoints every 250 epochs and generate 1,000 crystal structures from each checkpoint for comprehensive evaluation. All training dynamics plots use a 10-point moving average to reduce fluctuations inherent in the stochastic generation process. Note that the higher UN rates observed in this analysis (compared to the main paper's 10,000-structure evaluation) result from the smaller sample size, as uniqueness and novelty metrics naturally increase with fewer generated structures.

\subsubsection{Balance Score Validation Across Different Alpha Values}

\begin{figure}[t]
\centering
\begin{subfigure}{0.48\textwidth}
    \includegraphics[width=\textwidth]{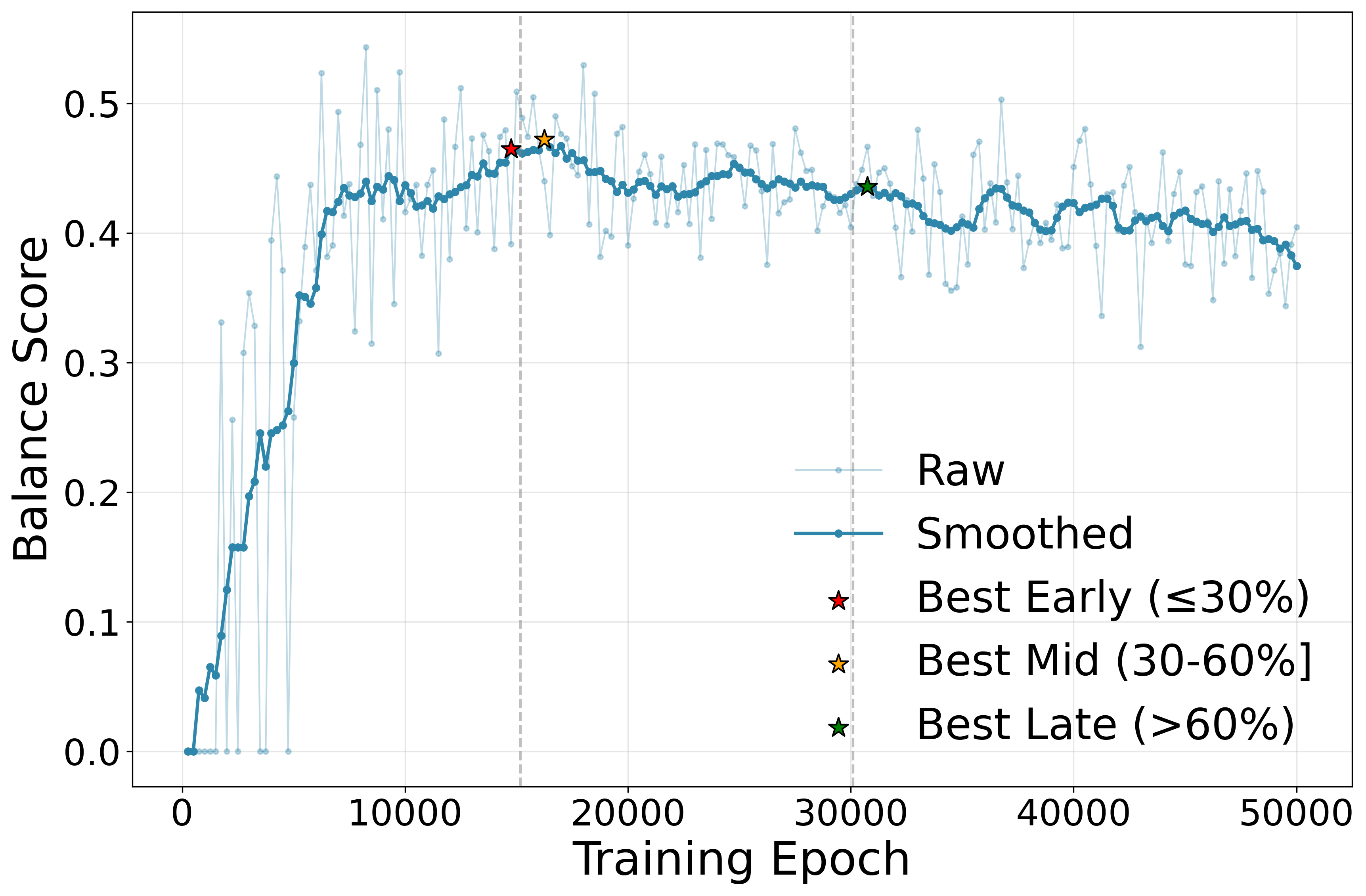}
    \caption{$\alpha = 1.0$}
\end{subfigure}
\hfill
\begin{subfigure}{0.48\textwidth}
    \includegraphics[width=\textwidth]{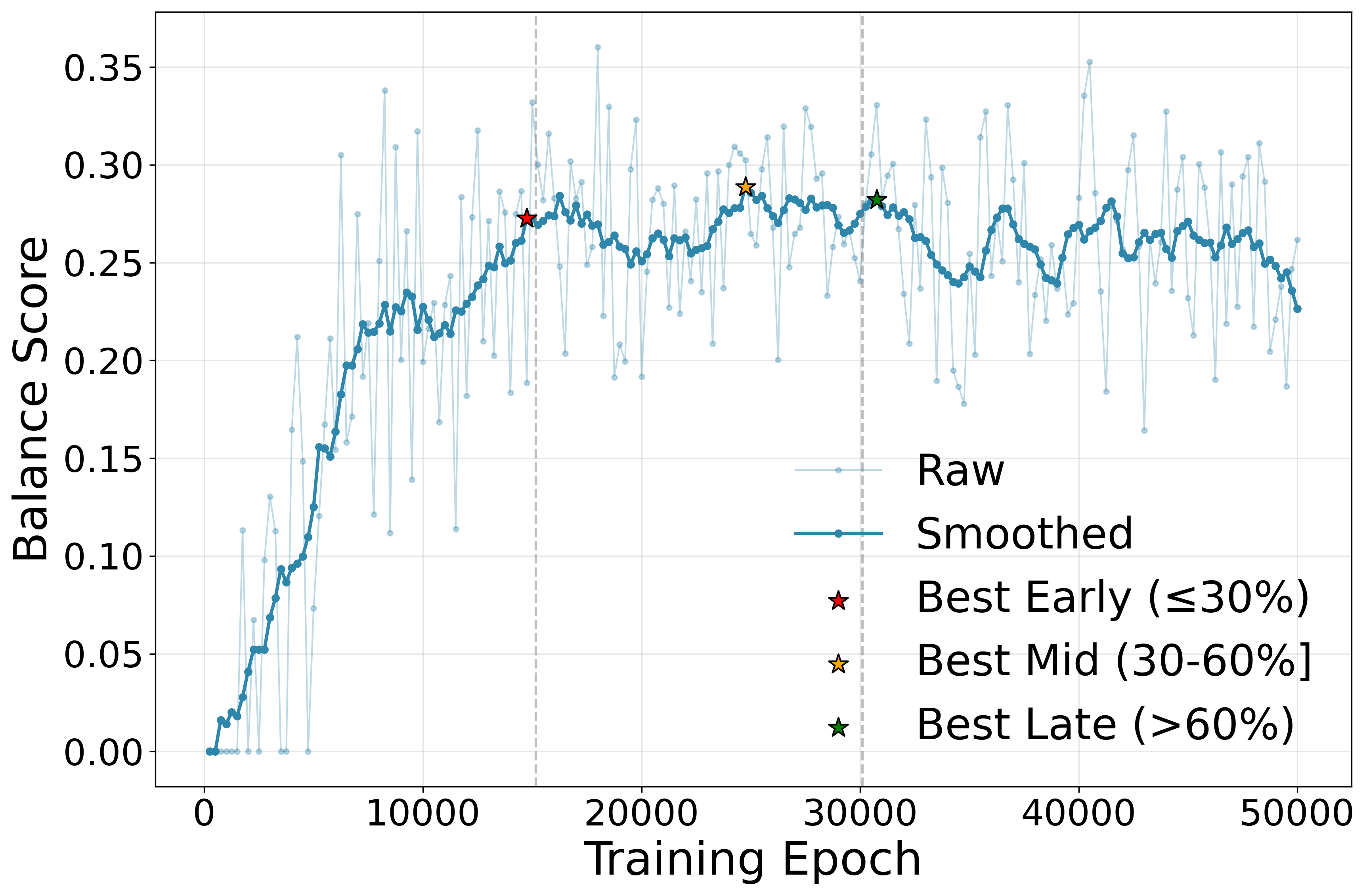}
    \caption{$\alpha = 2.0$}
\end{subfigure}
\caption{Balance Score evolution during training for different $\alpha$ values. Stars indicate best checkpoints selected from each training stage (early $\leq$30\%, mid 30-60\%, late $>$60\%).}
\label{fig:balance_score_comparison}
\end{figure}

Figure~\ref{fig:balance_score_comparison} presents Balance Score evolution for $\alpha = 1.0$ and $\alpha = 2.0$, demonstrating the impact of the quality-discovery trade-off parameter. Both curves show similar overall trends but differ in their checkpoint selection preferences.

The key difference emerges in the mid-stage selection (30-60\% of training): $\alpha = 1.0$ selects epoch 16,250, while $\alpha = 2.0$ selects epoch 24,750. This demonstrates that higher $\alpha$ values emphasize quality components more heavily, leading to selection of later checkpoints with improved structural and compositional validity. The early and late stage selections remain consistent across both values, indicating robust performance at training extremes.

\subsubsection{Training Dynamics and Quality-Discovery Trade-off}

\begin{figure}[t]
\centering
\begin{subfigure}{0.32\textwidth}
    \includegraphics[width=\textwidth]{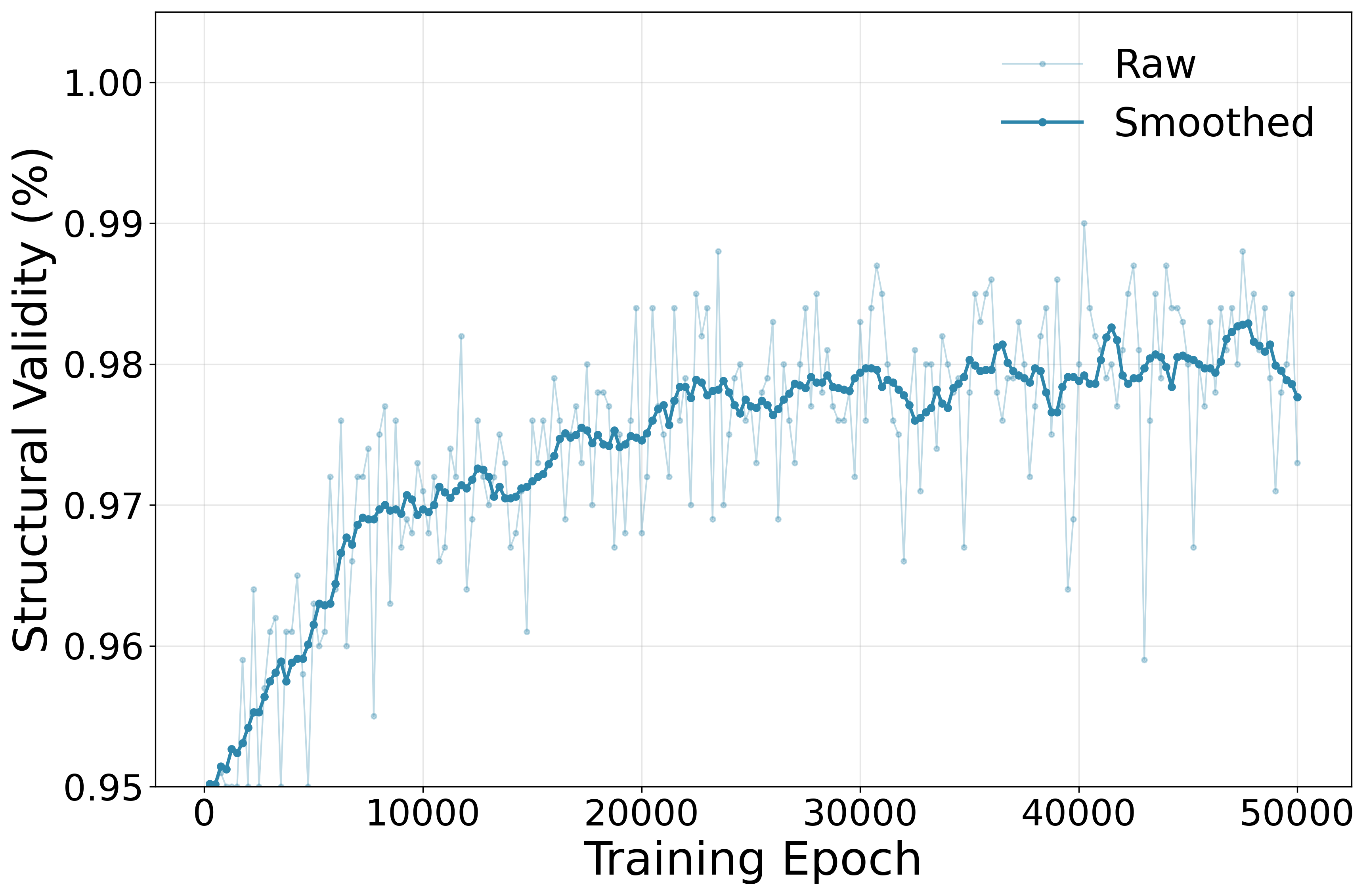}
    \caption{Structural Validity}
\end{subfigure}
\hfill
\begin{subfigure}{0.32\textwidth}
    \includegraphics[width=\textwidth]{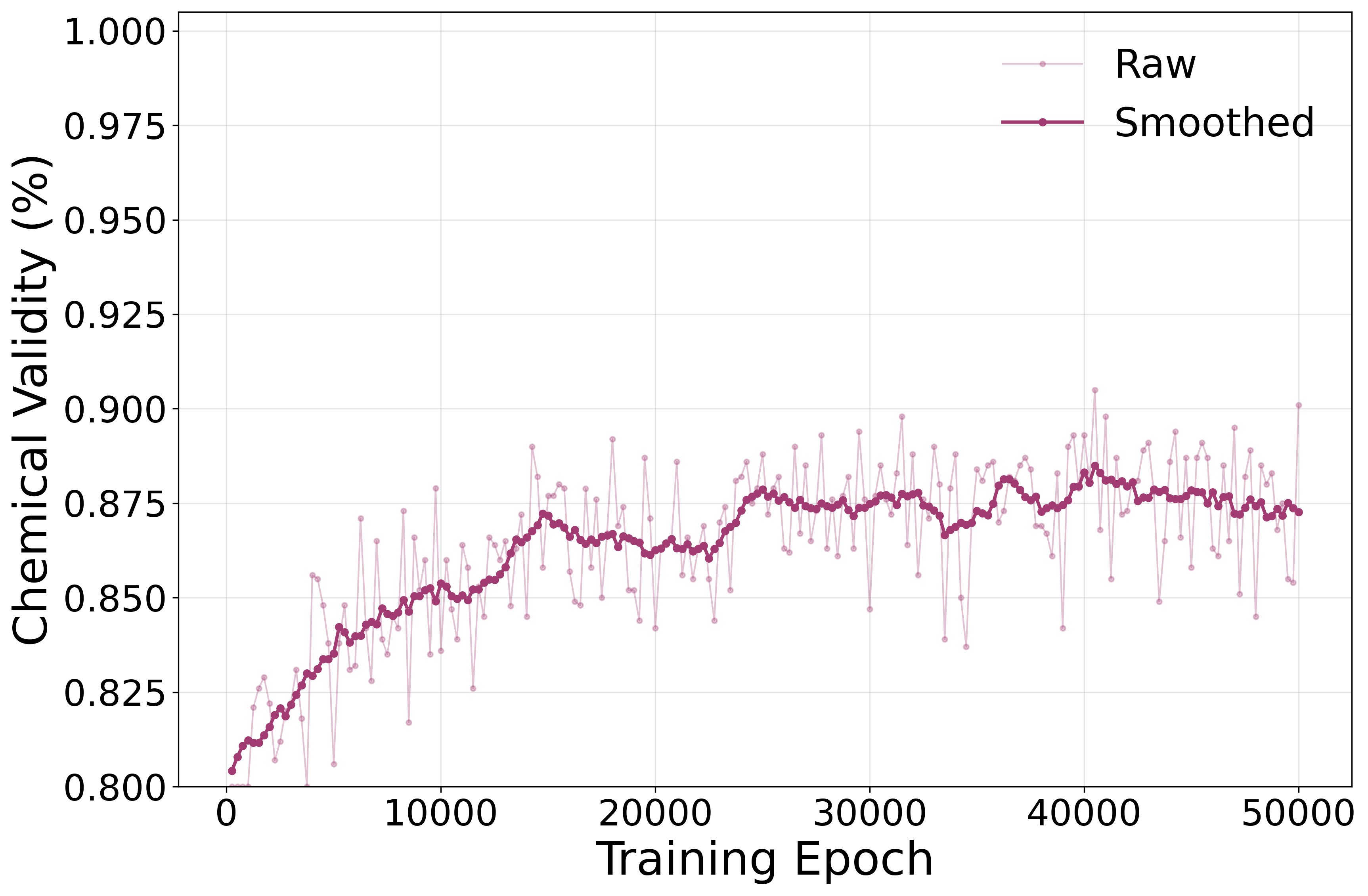}
    \caption{Chemical Validity}
\end{subfigure}
\hfill
\begin{subfigure}{0.32\textwidth}
    \includegraphics[width=\textwidth]{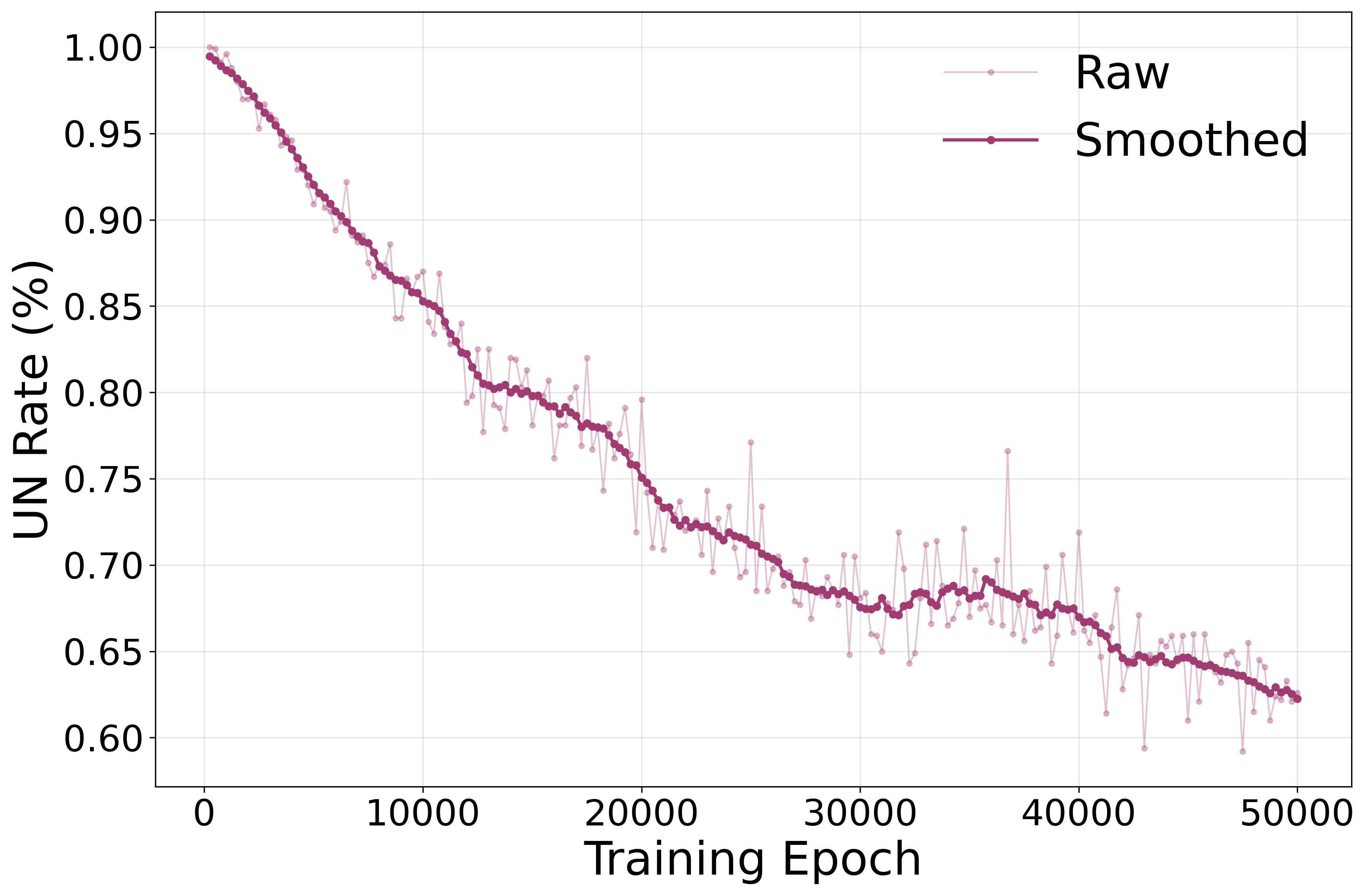}
    \caption{UN Rate}
\end{subfigure}
\caption{Training dynamics showing the quality-discovery trade-off. }
\label{fig:training_dynamics}
\end{figure}

Figure~\ref{fig:training_dynamics} illustrates the fundamental trade-off that motivates our Balance Score approach. As training progresses, we observe divergent trends in key metrics that traditional validation approaches fail to capture.

Structural validity demonstrates steady improvement from 95\% to 98\%, while chemical validity increases from 80\% to 88\%. However, the UN rate exhibits the opposite trend, declining from approximately 99.5\% to 62\%. This divergence highlights the critical limitation of traditional validity-based model selection: models that excel at reproducing training data patterns (high validity) may sacrifice the exploration necessary for materials discovery (low UN rate).

The substantial fluctuations observed in all metrics reflect the inherent stochasticity of diffusion model generation. These variations necessitate our 10-point smoothing window and underscore the importance of systematic checkpoint evaluation rather than relying on single-point assessments.

\subsubsection{Three-Stage Checkpoint Selection Rationale}

Our three-stage selection strategy addresses the temporal evolution of the quality-discovery trade-off. Early-stage models (epochs $\leq$15,000) exhibit high exploration capacity with UN rates exceeding 80\% but moderate validity scores. Late-stage models (epochs $>$30,000) achieve excellent validity metrics but reduced novelty generation. Mid-stage models represent optimal compromise points where both quality and discovery potential remain balanced.

This temporal analysis validates our multi-stage selection approach: no single training phase consistently produces optimal models for materials discovery. Instead, the best checkpoint emerges from systematic evaluation across all training stages using our Balance Score framework.

\subsection{Complete Energy Distribution Comparisons}

\subsubsection{Thermodynamic Favorability Analysis}

\begin{figure}[t]
\centering
\begin{subfigure}{0.48\textwidth}
    \includegraphics[width=\textwidth]{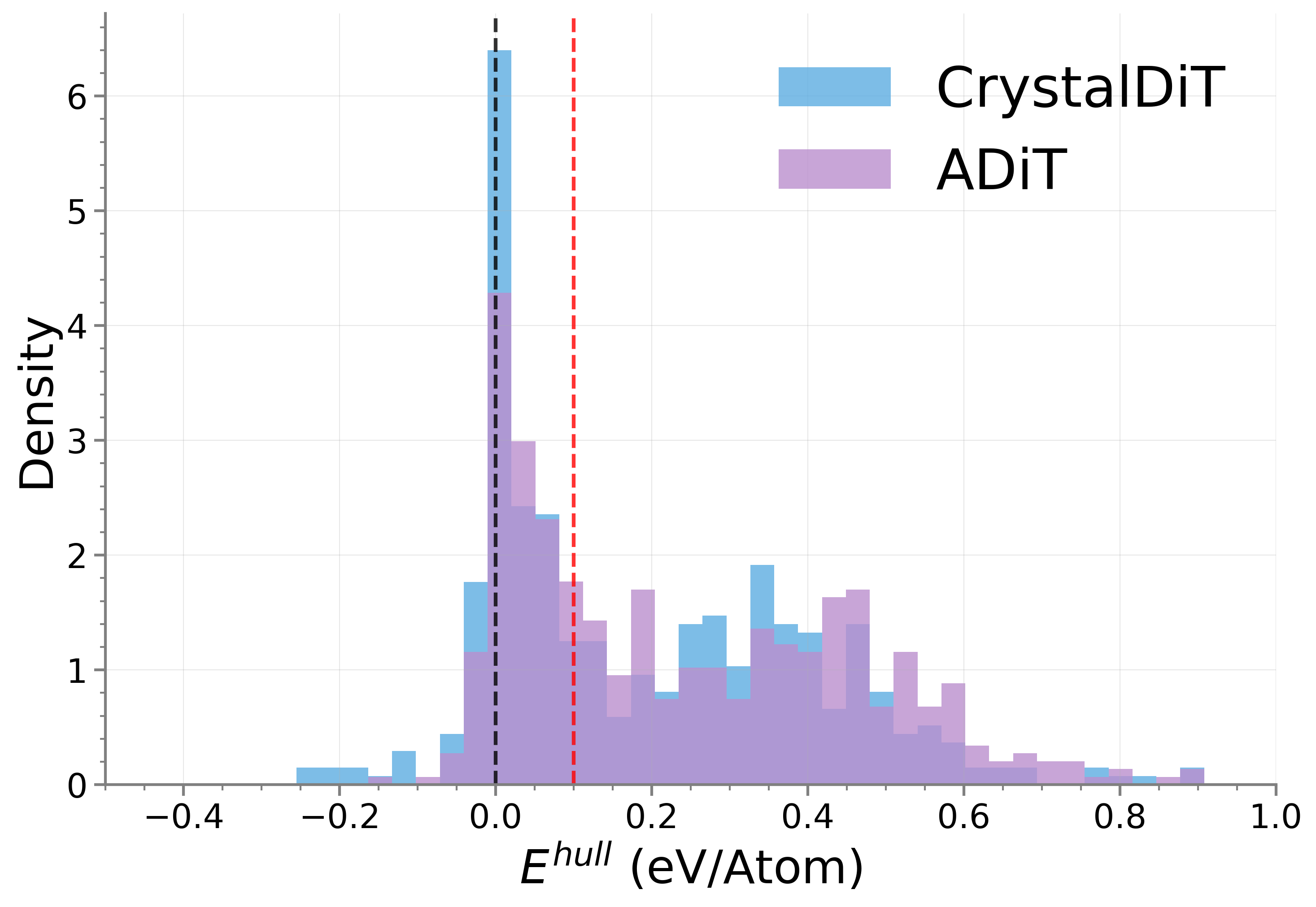}
    \caption{CrystalDiT vs ADiT}
\end{subfigure}
\hfill
\begin{subfigure}{0.48\textwidth}
    \includegraphics[width=\textwidth]{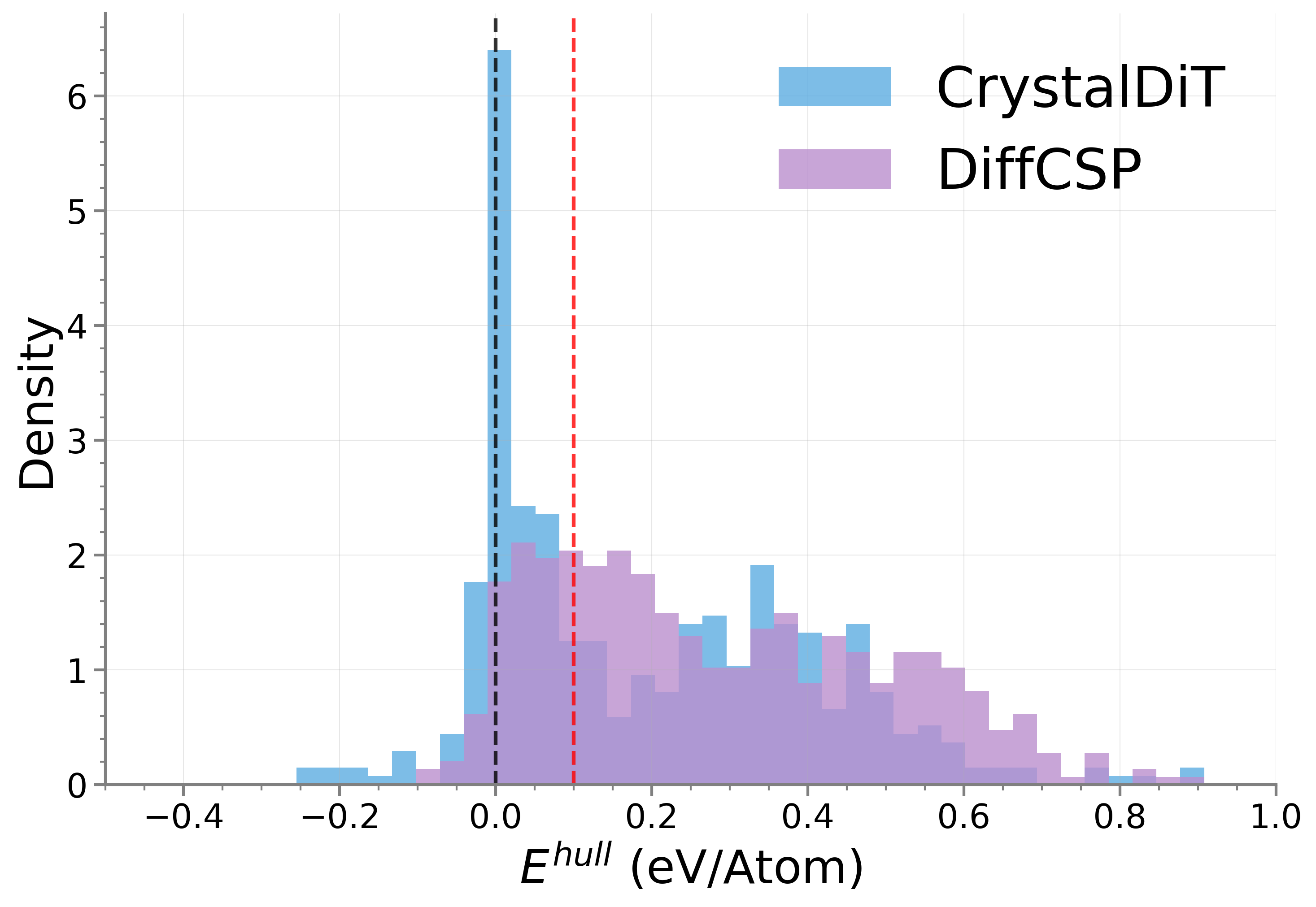}
    \caption{CrystalDiT vs DiffCSP}
\end{subfigure}

\begin{subfigure}{0.48\textwidth}
    \includegraphics[width=\textwidth]{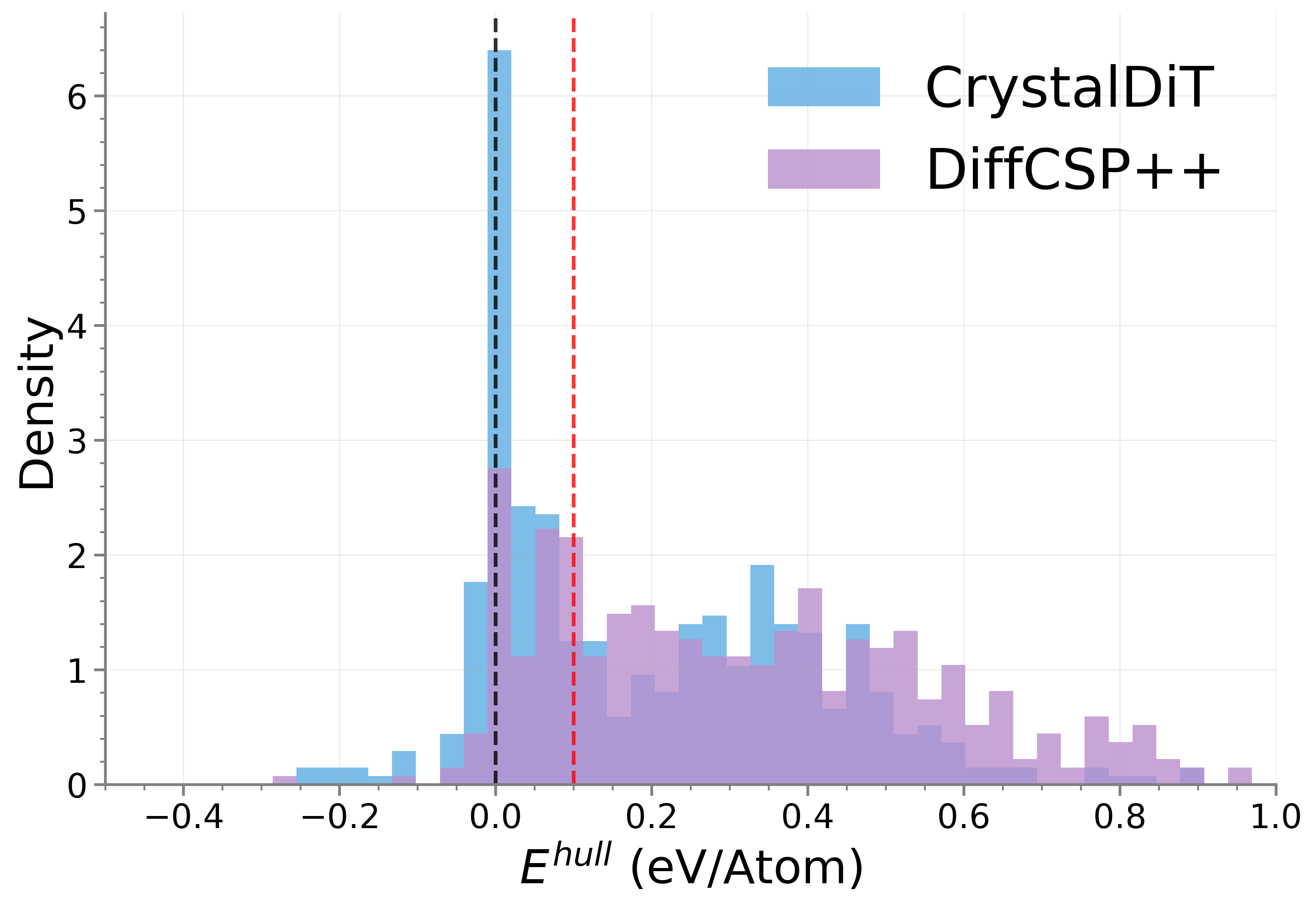}
    \caption{CrystalDiT vs DiffCSP++}
\end{subfigure}
\hfill
\begin{subfigure}{0.48\textwidth}
    \includegraphics[width=\textwidth]{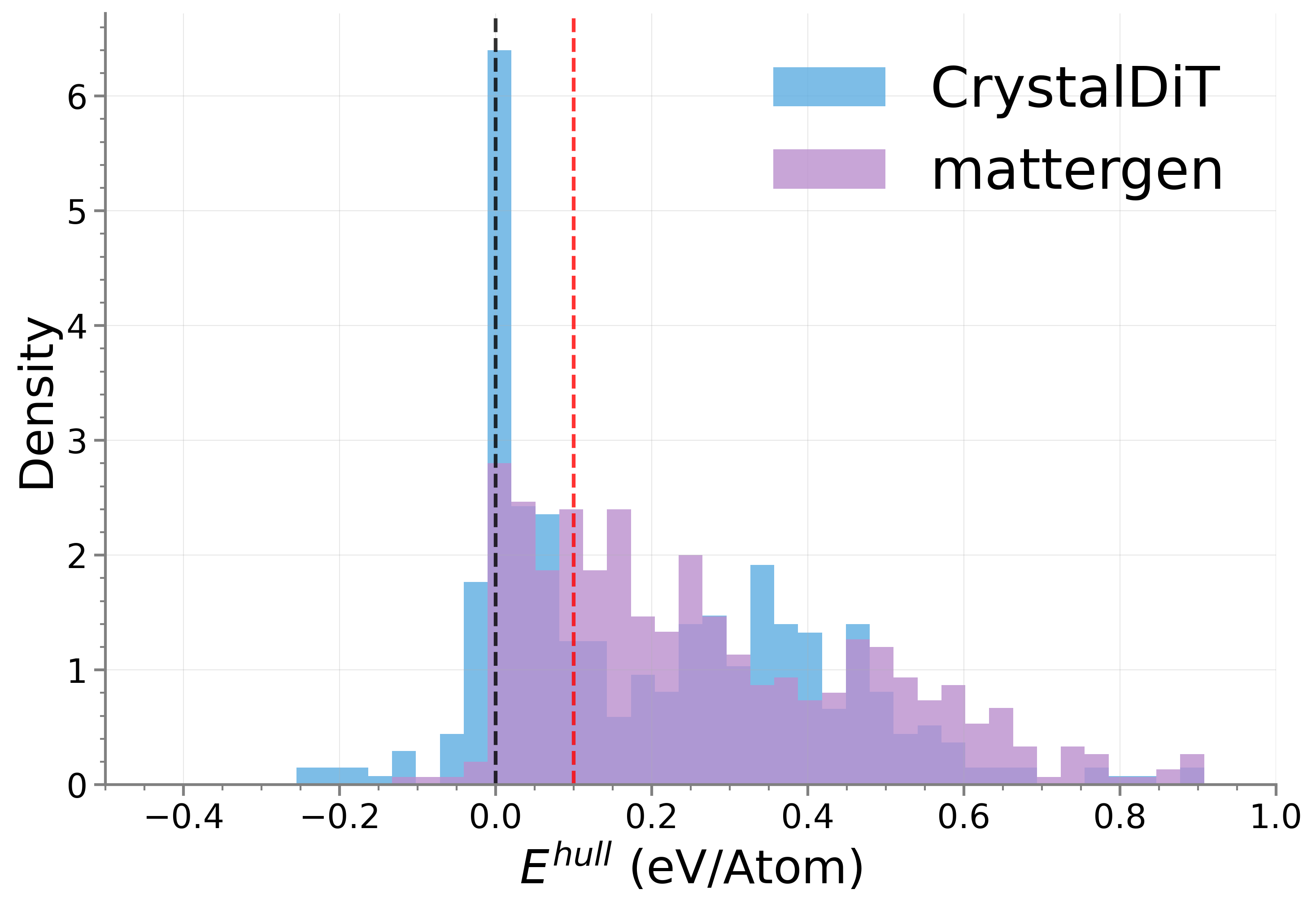}
    \caption{CrystalDiT vs MatterGen}
\end{subfigure}
\caption{Energy distribution comparisons between CrystalDiT and all baseline methods. Black dashed line: stability threshold ($E^{\text{hull}} = 0$); red dashed line: metastability threshold ($E^{\text{hull}} = 0.1$ eV/atom). CrystalDiT consistently shows higher density in stable and metastable regions.}
\label{fig:energy_distributions_complete}
\end{figure}

To complement the FlowMM comparison presented in the main paper, Figure~\ref{fig:energy_distributions_complete} provides comprehensive energy distribution analysis across all baseline methods. These distributions reveal CrystalDiT's superior ability to generate thermodynamically favorable crystal structures.

Across all comparisons, CrystalDiT demonstrates consistent advantages in thermodynamic favorability:

\textbf{Enhanced Stability Generation}: In the thermodynamically stable region ($E^{\text{hull}} < 0$), CrystalDiT exhibits pronounced density peaks significantly higher than all baseline methods. This indicates superior capability for generating energetically favorable crystal structures.

\textbf{Concentrated Energy Distributions}: CrystalDiT produces sharper peaks near 0 eV/atom, suggesting generated structures cluster closer to thermodynamic equilibrium compared to the broader, more diffuse distributions of baseline methods.

\textbf{Reduced High-Energy Structures}: In the high-energy region ($E^{\text{hull}} > 0.1$ eV/atom), CrystalDiT generates substantially fewer structures, indicating better avoidance of thermodynamically unfavorable configurations.

These energy distribution patterns provide crucial validation that CrystalDiT's simplified architecture not only generates more unique and novel structures but ensures these structures exhibit superior thermodynamic viability for practical materials applications. The consistent performance across all baseline comparisons reinforces the effectiveness of our unified attention mechanism for learning physically meaningful crystal generation patterns.

\subsection{Scaling to Larger Crystal Structures}

To evaluate CrystalDiT's generalization capability beyond MP-20's 20-atom limit, we conduct experiments on MPTS-52, a more challenging dataset containing crystal structures with up to 52 atoms. We train CrystalDiT (Simple) using identical hyperparameters, with only the maximum atom count increased from 20 to 52.

\begin{table}[htbp]
\centering
\small
\begin{tabular}{l|c|cc}
\toprule
\textbf{Dataset} & \textbf{UN Rate (\%)} & \textbf{SUN (\%)} & \textbf{MSUN (\%)} \\
\midrule
MP-20   & 63.28 & 8.78 & 25.90 \\
MPTS-52 & 61.45 & 6.73 & 20.19 \\
\midrule
Degradation & -1.83 & -2.05 & -5.71 \\
\bottomrule
\end{tabular}
\caption{CrystalDiT performance comparison between MP-20 and MPTS-52 datasets.}
\label{tab:mpts52_results}
\end{table}

CrystalDiT achieves 6.73\% SUN rate on MPTS-52, with only 2.05\% absolute decrease compared to MP-20 performance despite the 2.6× increase in maximum structure size. This modest degradation demonstrates that our simplified architecture and periodic table representation effectively capture fundamental chemical principles that transfer across different structural complexities, validating the practical utility of our approach for discovering larger, more complex materials.

\section{Visualizations}

\subsection{Generated Crystal Structures}

\begin{figure*}[htbp]
\centering
\begin{subfigure}[b]{0.42\textwidth}
    \centering
    \includegraphics[width=\textwidth]{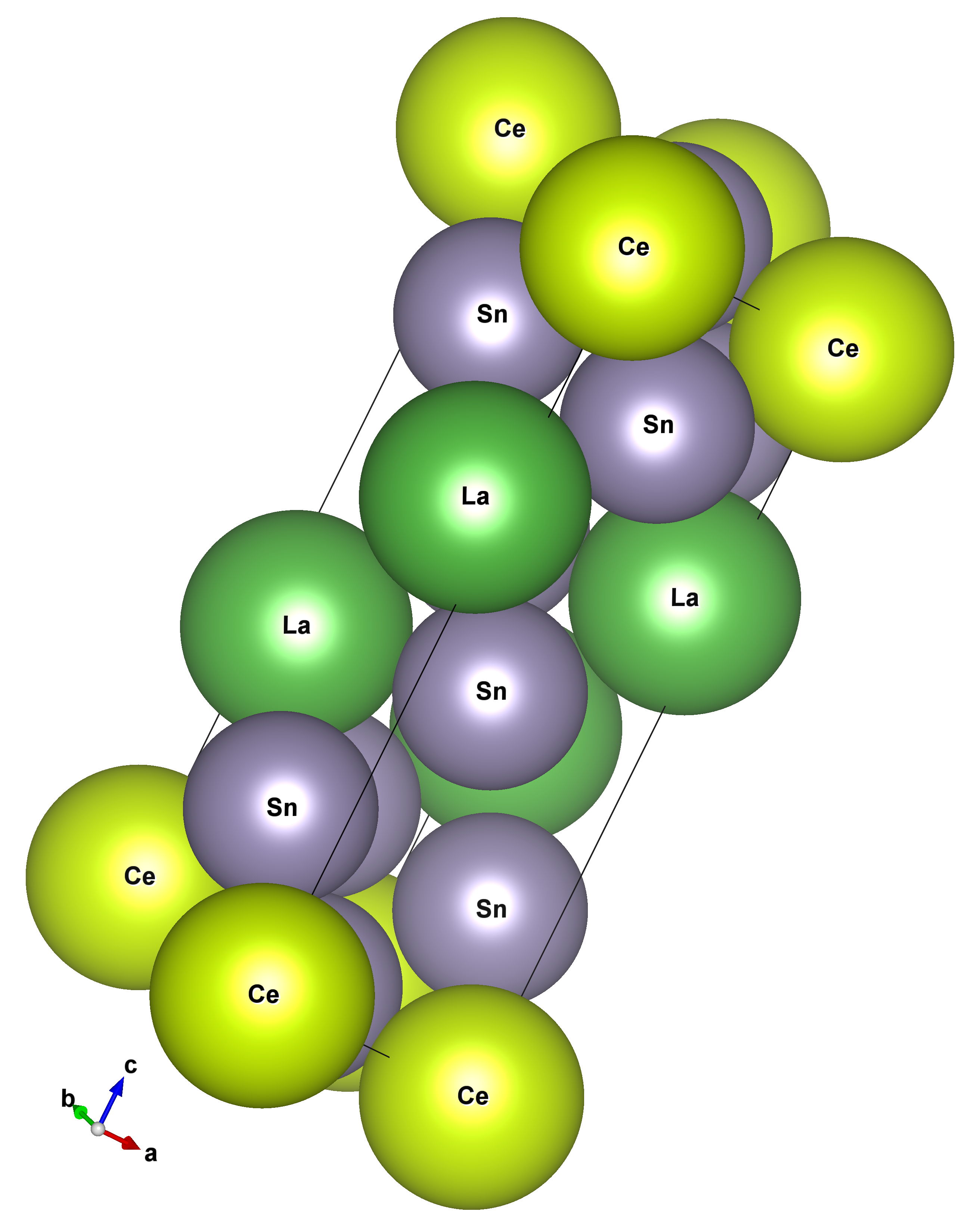}
    \caption{La$_2$CeSn$_7$ \\ Space Group: Cmmm}
    \label{fig:La2Ce1Sn7}
\end{subfigure}
\hfill
\begin{subfigure}[b]{0.42\textwidth}
    \centering
    \includegraphics[width=\textwidth]{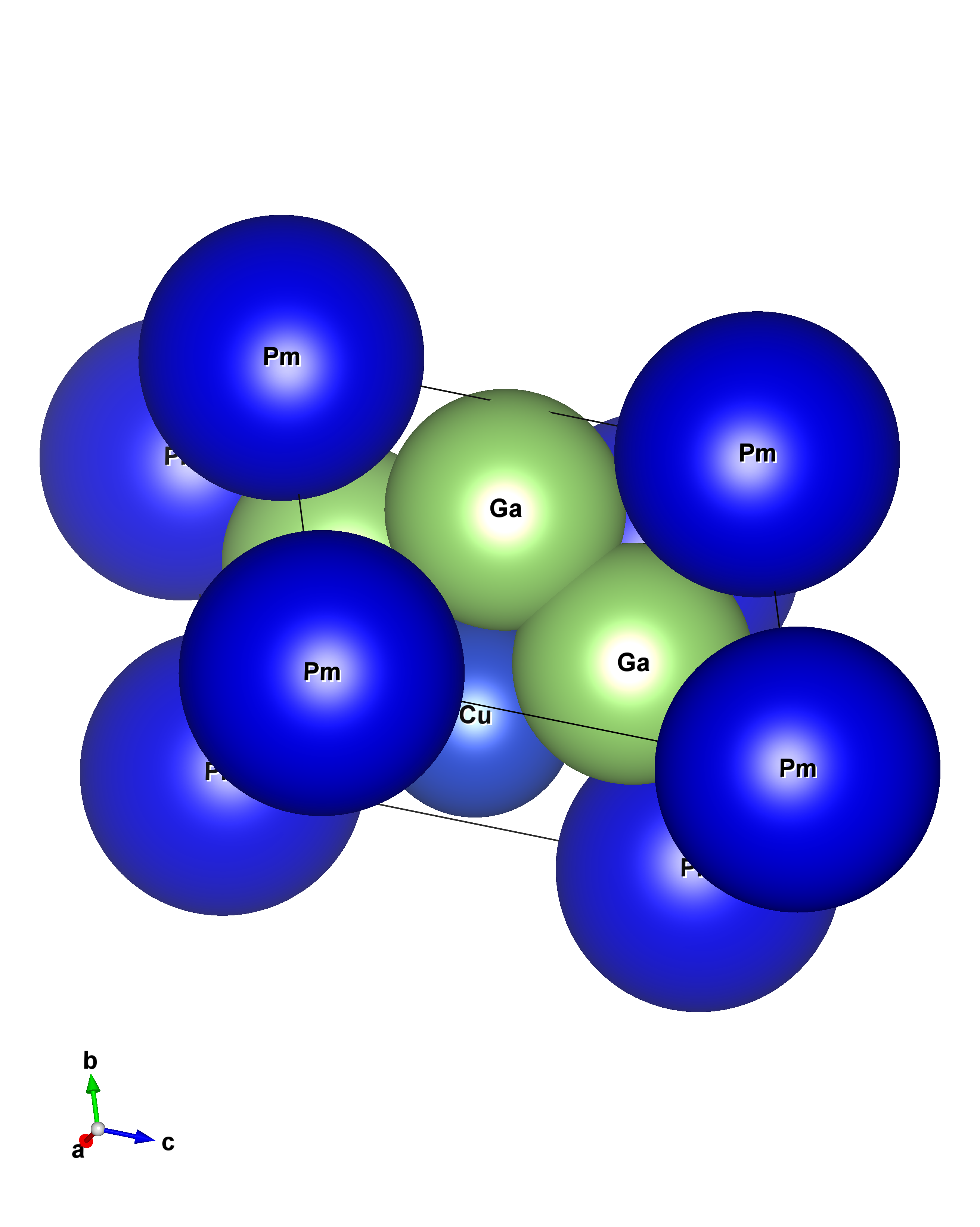}
    \caption{PmGa$_3$Cu \\ Space Group: I-4m2}
    \label{fig:Pm1Ga3Cu1}
\end{subfigure}

\vspace{0.5cm}

\begin{subfigure}[b]{0.42\textwidth}
    \centering
    \includegraphics[width=\textwidth]{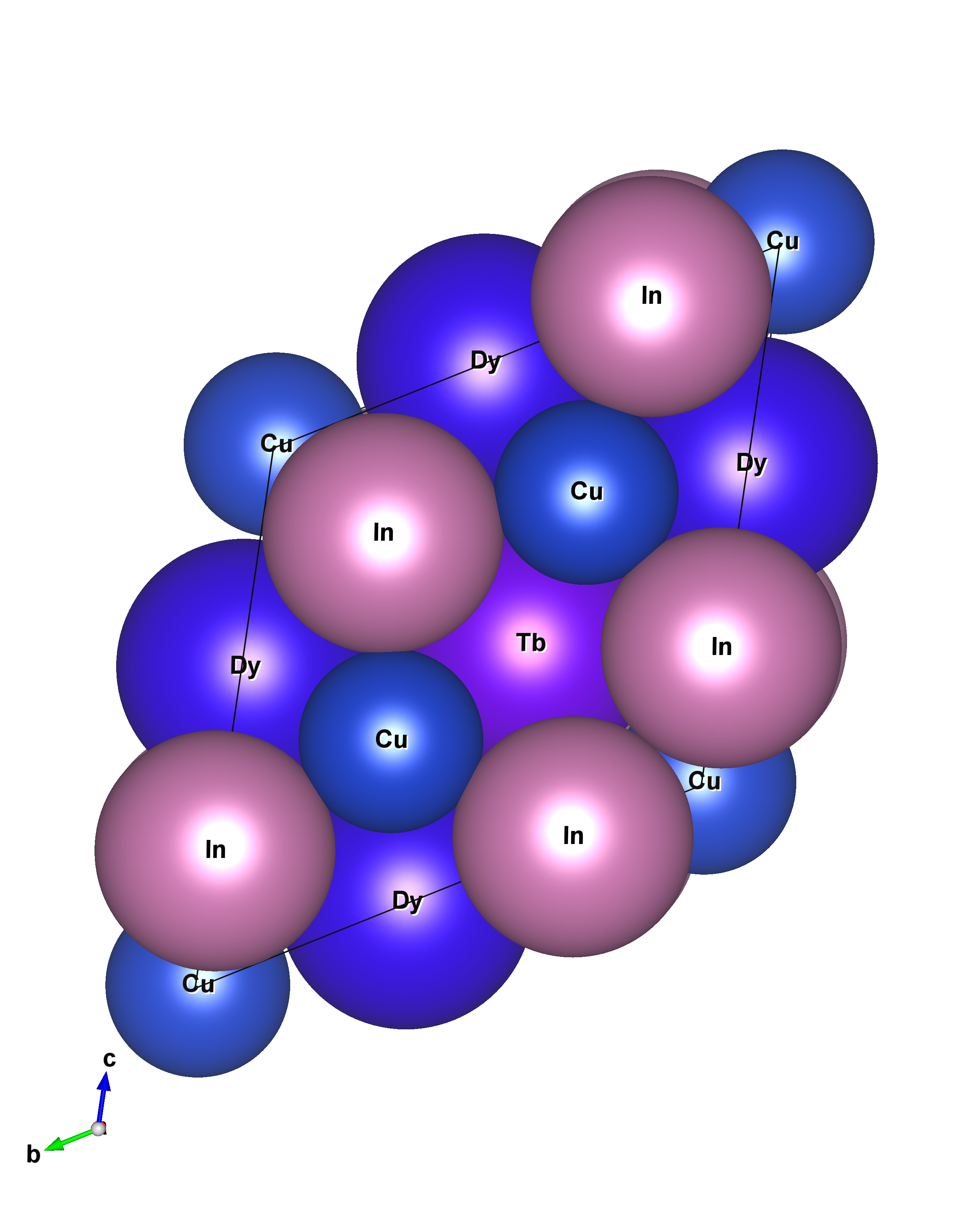}
    \caption{TbDy$_2$In$_3$Cu$_3$ \\ Space Group: Amm2}
    \label{fig:Tb1Dy2In3Cu3}
\end{subfigure}
\hfill
\begin{subfigure}[b]{0.42\textwidth}
    \centering
    \includegraphics[width=\textwidth]{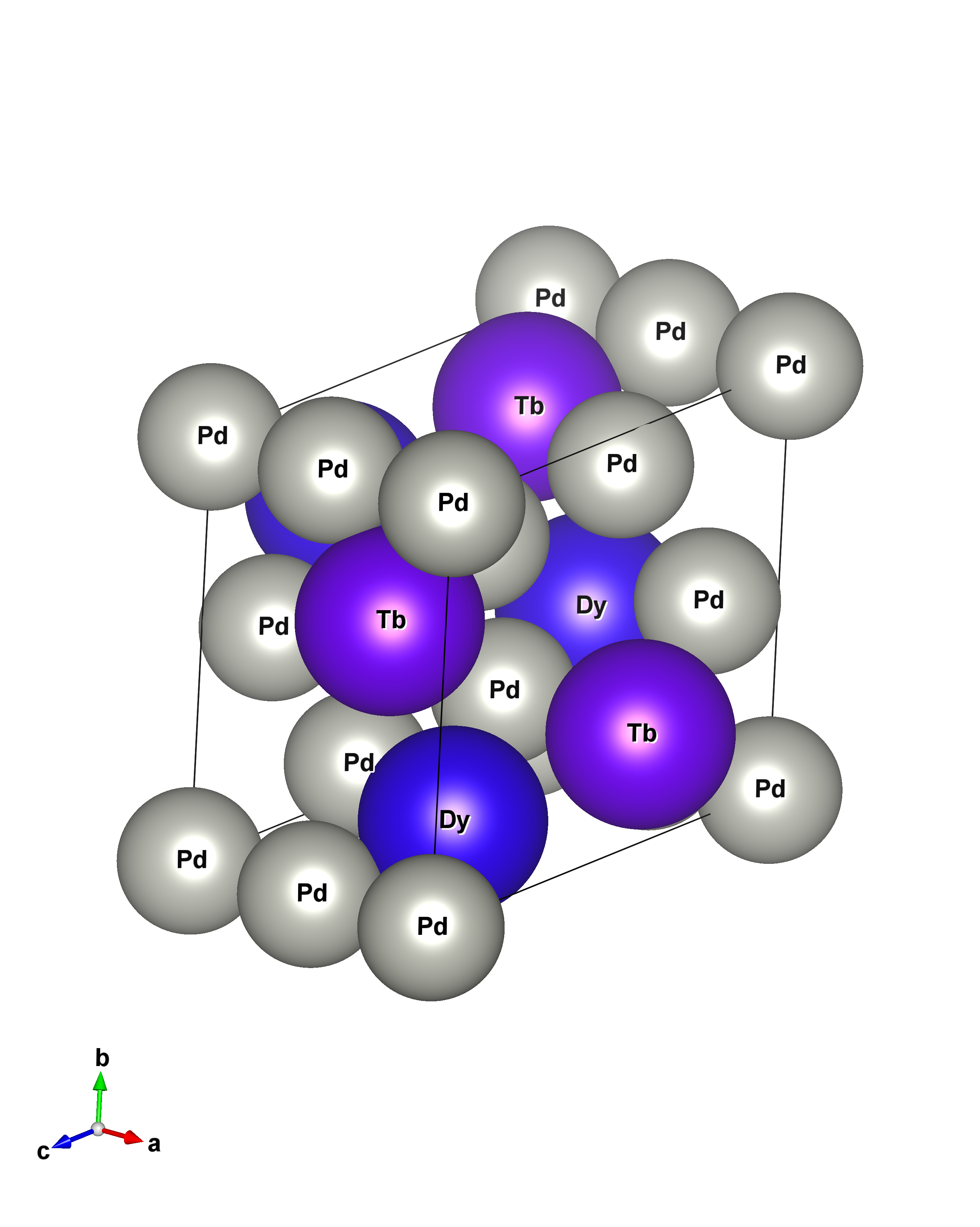}
    \caption{Tb$_3$Dy$_3$Pd$_8$ \\ Space Group: P1}
    \label{fig:Tb3Dy3Pd8}
\end{subfigure}

\caption{Representative crystal structures generated by CrystalDiT that satisfy the SUN criteria (Stable, Unique, Novel). Each structure demonstrates successful generation of chemically reasonable and thermodynamically stable materials across diverse chemical systems and space groups. The structures showcase CrystalDiT's ability to generate (a) rare earth stannides with orthorhombic symmetry, (b) rare earth intermetallics with tetragonal symmetry, (c) complex rare earth-transition metal compounds with orthorhombic polar symmetry, and (d) mixed rare earth-platinum group metal alloys with triclinic symmetry. }
\label{fig:generated_structures}
\end{figure*}

Figure~\ref{fig:generated_structures} showcases representative crystal structures generated by CrystalDiT that achieve both high uniqueness/novelty and thermodynamic stability. These structures demonstrate the model's ability to discover diverse crystal chemistries while maintaining realistic atomic arrangements and space group symmetries. Each structure represents a successful example from our SUN (Stable, Unique, Novel) evaluation, confirming that CrystalDiT can generate materials with genuine discovery potential.

The visualization reveals several key insights: (1) CrystalDiT generates structures across diverse chemical systems, from rare earth compounds to transition metal alloys; (2) the generated structures exhibit proper coordination environments and realistic bond lengths; (3) the model successfully captures various space group symmetries, from high-symmetry tetragonal systems to low-symmetry triclinic structures; and (4) the chemical compositions follow reasonable stoichiometric ratios consistent with materials science principles.

The structural diversity demonstrated in these examples validates our approach's effectiveness for materials discovery. The La$_2$CeSn$_7$ structure represents a rare earth stannide with complex layered arrangement typical of RE$_2$TSn$_7$ compounds, while PmGa$_3$Cu exhibits the tetragonal structure common in rare earth-gallium-copper systems. The TbDy$_2$In$_3$Cu$_3$ structure showcases CrystalDiT's ability to generate mixed rare earth compounds with multiple transition metals, and Tb$_3$Dy$_3$Pd$_8$ demonstrates successful generation of complex platinum group metal alloys.

These visualizations confirm that CrystalDiT not only achieves high SUN rates quantitatively but also generates structures that are chemically meaningful and structurally realistic, making them viable candidates for experimental synthesis and further materials property exploration.

\end{document}